\documentclass[10pt,journal]{IEEEtran}
\usepackage[pdftex]{graphicx}
\DeclareGraphicsExtensions{.pdf,.jpg,.png}
\usepackage{cite}
\usepackage[cmex10]{amsmath}
\usepackage{amssymb}
\usepackage{graphicx}
\usepackage{cases}
\usepackage{mdwmath}
\usepackage{mdwtab}
\usepackage{array}
\usepackage{url}
\usepackage[ruled,vlined]{algorithm2e}
\usepackage{booktabs}
\usepackage{threeparttable}
\usepackage{multirow}
\usepackage{color}
\usepackage{caption}
\usepackage{algorithmic}
\usepackage{amsfonts,makecell}

\newcommand{\myPara}[1]{\vspace{.05in}\noindent\textbf{#1}}

\newcommand{\mb}[1]{\mathbb{#1}}

\newcommand{\bm}[1]{\mbox{\boldmath{$#1$}}}
\newcommand{\mr}[1]{\mathrm{#1}}

\graphicspath{{./fig/}}
\usepackage{color}

\ifCLASSOPTIONcompsoc
\usepackage[caption=false,font=footnotesize,labelfont=sf,textfont=sf]{subfig}
\else
\usepackage[caption=false,font=footnotesize]{subfig}
\fi

\usepackage{colortbl}
\usepackage{xcolor}
\usepackage{xpatch}

\begin{document}
	
	\title{Trustable Co-label Learning from Multiple Noisy Annotators}
	\author{Shikun Li, Tongliang Liu,~\IEEEmembership{Senior Member,~IEEE},
Jiyong Tan,
		Dan Zeng,~\IEEEmembership{Senior Member,~IEEE} and
		Shiming Ge,~\IEEEmembership{Senior Member,~IEEE}
		\thanks{S. Li and S. Ge are with the Institute of Information Engineering, Chinese Academy of Sciences, Beijing 100095, China, and with School of Cyber Security at University of Chinese Academy of Sciences, Beijing 100049, China. Email:~\{lishikun,geshiming\}@iie.ac.cn}
		\thanks{T. Liu is with the Trustworthy Machine Learning Lab, the University of Sydney, 6 Cleveland St, Darlington, NSW 2008, Australia. Email:~tongliang.liu@sydney.edu.au}
\thanks{J. Tan is with the AISONO AIR Lab, Shenzhen, and with the Harbin Institute of Technology. Email:~scutjy2015@163.com}
		\thanks{D.~Zeng is with the Department of Communication Engineering, Shanghai University, Shanghai 200040, China. E-mail:~dzeng@shu.edu.cn}
		\thanks{S. Ge is the corresponding author. E-mail:~geshiming@iie.ac.cn}
	}
	

	\maketitle

\begin{abstract}
Supervised deep learning depends on massive accurately annotated examples, which is usually impractical in many real-world scenarios. A typical alternative is learning from multiple noisy annotators. Numerous earlier works assume that all labels are noisy, while it is usually the case that a few trusted samples with clean labels are available. This raises the following important question: how can we effectively use a small amount of trusted data to facilitate robust classifier learning from multiple annotators? This paper proposes a data-efficient approach, called \emph{Trustable Co-label Learning} (TCL), to learn deep classifiers from multiple noisy annotators when a small set of trusted data is available. This approach follows the coupled-view learning manner, which jointly learns the data classifier and the label aggregator. It effectively uses trusted data as a guide to generate trustable soft labels (termed co-labels). A co-label learning can then be performed by alternately reannotating the pseudo labels and refining the classifiers. In addition, we further improve TCL for a special complete data case, where each instance is labeled by all annotators and the label aggregator is represented by multilayer neural networks to enhance model capacity. Extensive experiments on synthetic and real datasets clearly demonstrate the effectiveness and robustness of the proposed approach. Source code is available at \textcolor{magenta}{https://github.com/ShikunLi/TCL}.
\end{abstract}

	
	\IEEEpeerreviewmaketitle
	
	\section{Introduction}
	
	\IEEEPARstart{R}{ecent} deep learning has achieved state-of-the-art results in various classification tasks, e.g., image recognition~\cite{he2016deep}, object detection~\cite{ren2015faster} and text matching~\cite{ChenHNLLWX18}. These advances are largely due to the availability of large-scale cleanly annotated datasets and effective model learning algorithms. However, massive clean annotations are very difficult to collect in many real-world scenarios, e.g., video surveillance in the wild~\cite{Sun2018DPPDL}, medical data analysis~\cite{WangLLXRZMLHZ20} and webly supervised multimedia understanding~\cite{YaoZSHXT17,WangWJ14,ChaudharyGPC20,9113752,abs-1708-02862}. Noisy labels are usually easier to acquire in these scenarios. Since the capacity of deep networks is so high that they can totally memorize noisy labels~\cite{Zhang2017iclr}, designing deep methods robust to noisy labels is challenging~\cite{Liu2016TPAMI,cheng2019learning,XiaLW00NS19}.

Typically, learning from multiple noisy annotators (e.g., different non-expert persons, weak models, weak discriminant rules, or other automatic labeling sources) provides an alternative way to lessen label noise~\cite{DengDSLL009,BielG13,ServajeanJSCP17}. The common practice is to aggregate multiple weak labels for one instance to obtain a more reliable label and then learn with the aggregated labels. Following this line, many label aggregation methods have been proposed, e.g., majority vote~\cite{WhitehillRWBM09}, participant-mine voting~\cite{LiLGZFH14}, Dawid-Skene estimator~\cite{Dawid1979Maximum} and Bayesian classifier combination~\cite{KimG12}. In addition, a promising direction is to introduce the instance feature to help decrease the influence of label noise ~\cite{Raykar2010JMLR,ZhangSLW18,Albarqouni2016TMI,Rodrigues2017TPAMI,YinL0020}.
Recently, some studies aimed in this direction jointly learn a deep classifier and a label aggregator~\cite{CaoXKW19,Khetan2018iclr,Li2020aaai}, which provides a flexible and general coupled-view deep learning manner to address the issue.
However, these works assume that all labels are noisy, while it is usually the case that a few trusted examples with clean labels are available. This means that if we effectively introduce such trusted data into training,
it will enable substantial label corruption robustness performance gains. Then, this raises the following important question: \emph{how can we effectively use the trusted data to facilitate robust classifier learning from multiple annotators?}

\begin{figure}
		\centering
		\includegraphics[width=1.0\linewidth]{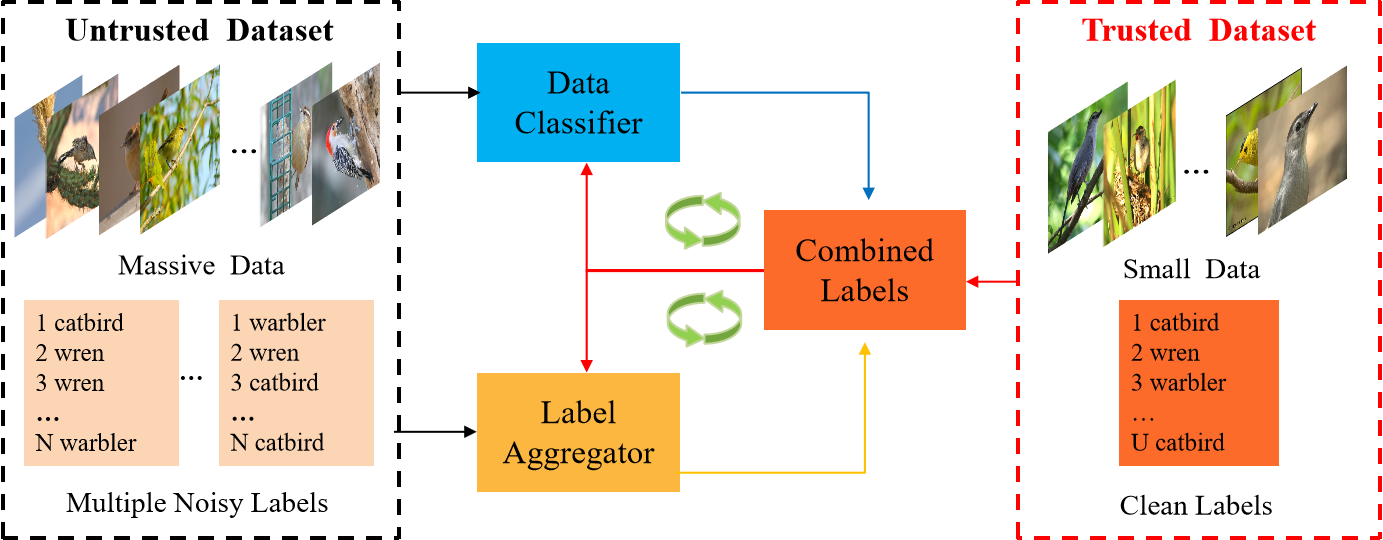}
\caption{Our approach alternately refines classifiers and reannotates co-labels with the guidance of a small, trusted dataset. }
		\label{fig:motivation}
\end{figure}
An intuitive way to utilize the small, trusted dataset is fine-tuning the pretrained models~\cite{yu2010roles}, where the trusted data usually do not play a role in the pretraining process and the small amount of data limits its performance gains. Thus, we propose to improve the model performance during the whole learning process with the help of the small, trusted dataset.
As illustrated in Fig. \ref{fig:motivation}, our idea is to jointly model the data classifier by deep neural networks and the label aggregator by a Naive Bayes classifier in a coupled-view learning fashion. During training, our approach alternately refines classifiers and combines the predictions into trustable soft labels (termed co-labels) with the guidance of the small, trusted dataset.
In the iterative learning process, to achieve a trustable prediction combination, we first utilize the trusted data to calibrate the data classifier, which makes the confidences from the deep classifier reflect the ground truth correctness likelihood~\cite{GuoPSW17}. 
Then, based on the conditionally independent probability assumption, we combine the label probabilities from both classifiers into a joint probability, which is regarded as a trustable soft label distribution for robust classifier learning in the next iteration. After the alternate optimization, our approach reinitializes the data classifier, which is then retrained on the union of reannotated untrusted and trusted data. In this way, the small, trusted dataset provides both its original clean target for classifier learning and extra guidance for the reliable labeling of untrusted data, leading to data-efficient learning.
	Our approach is named \emph{Trustable Co-label Learning} (TCL) since it effectively makes use of a small, trusted dataset to progressively obtain more trustable labels by combining the data classifier and the label aggregator.
	
	In addition, we design a variant method of TCL for a special complete data case, where all instances are labeled by every annotator. Note that this kind of data can be collected from various automatic labeling sources, e.g., weak classifiers~\cite{Li2017iccv}, user behaviors~\cite{9337209} and search engines~\cite{DengJTGL14}. To achieve reliable co-label updates, we similarly obtain well-calibrated deep networks and combine predictions via the trusted data. Different from the original TCL, we model both the data classifier and the label aggregator by multilayer neural networks. It is natural that the performance of label aggregator benefits from the high modeling capacity of deep networks.
	
	To verify the effectiveness and robustness of the proposed approach, {we conduct experiments on synthetic and real datasets under two learning settings, i.e., learning from non-expert humans (sparse data case\footnote{In this paper, the sparse data case means each annotator only labels part of training data, and the complete data case means all training samples are labeled by every annotator.}) and learning from auto-labeling sources (complete data case). The results clearly show that our approach can substantially combat the negative impact of label noise by using a small, trusted dataset effectively.
	
	The main contributions of this paper are summarized in three aspects:
1) We propose a data-efficient approach, called \emph{Trustable Co-label Learning} 
(TCL)
, to learn a robust classifier from multiple noisy annotators when a small, trusted dataset is available. It effectively uses trusted data to combine the predicted distributions into the trustable label distributions.
2) We further improve TCL for a special complete data case, where all instances are labeled by every annotator, and the label aggregator is represented by multilayer neural networks to enhance modeling ability.
3) We conduct experiments on synthetic and real datasets, which clearly demonstrate that our approach outperforms state-of-the-art approaches in terms of effectiveness and robustness.
	
	\section{Related Works}
	We briefly review the related works from three aspects in this section, including learning with multiple noisy labels, multi-view methods and hybrid methods.
	\subsection{Learning with Multiple Noisy Labels}
	
	When multiple noisy annotators are available for each instance, one basic direction is to infer true labels from multiple noisy labels and then learn a data classifier with those inferred labels. Probabilistic generative methods and discriminative methods are mainly two strands to address such label aggregation problems. Generative methods generally build a probabilistic model to generate noisy observations conditioned on unknown true labels and some behavior assumptions, e.g., the Dawid-Skene estimator~\cite{Dawid1979Maximum}, the minimax entropy estimator~\cite{ZhouPBM12}, {Bayesian classifier combination~\cite{KimG12}, structure learning~\cite{BachHRR17} and their variants~\cite{ZhouLPM14,WhitehillRWBM09,SimpsonRPS13,BiWKT14,Kurve0K15,VarmaSHRR19}.} In contrast, discriminative methods do not model the observations but directly identify the true labels via aggregation rules. Between them, the simplest but effective method is majority voting~\cite{IpeirotisPSW14}, which Naively assumes that all annotators are equally reliable. Advanced approaches take different reliabilities of workers or instances into consideration, including weighted majority voting~\cite{Li014,LiLGZFH14}, graph modeling~\cite{ParameswaranSGPW11}, max-margin majority voting~\cite{TianZ15}, tensor factorization methods~\cite{KargerOS11,DalviDKR13,ZhouH16}, etc.
	
	Recently, a promising direction has been to introduce instance features to help decrease the influence of label noise, especially to jointly learn data classifiers.
	Many works in this direction propose an 
Expectation Maximization (EM)
 algorithm for jointly learning the levels of expertise of different annotators and the parameters of a classifier~\cite{Raykar2010JMLR,Albarqouni2016TMI,Rodrigues2017TPAMI,Khetan2018iclr}. Rodrigues et al.~\cite{Rodrigues2018aaai} propose adding a crowd layer to the output of a common network, and the layer adjusts the gradients coming from the labels of annotators. DoctorNet~\cite{Guan2018aaai} learns different models for every annotator, and the whole output is weighted integration of multiple models' predictions. Chu et al.~\cite{Chu0W21} also add a crowd layer but decompose the confusion matrix into a commonly shared confusion matrix and an individual confusion matrix. Cao et al.~\cite{CaoXKW19} provide an information theoretic method that interprets the joint learning-from-multiple-annotators problem as a coupled-view problem. Li et al.~\cite{Li2020aaai} introduce a coupled-view method with several robust learning schemes to address it. Generally, these works assume that all labels are noisy and may be suboptimal when a small, trusted dataset is available. Therefore, we effectively use the trusted data in a coupled-view learning manner to improve the learning performance.

\subsection{Multi-view Methods}
Multi-view methods mainly include co-training style algorithms, co-regularization style algorithms, multi-kernel learning algorithms and other types. Co-training style algorithms~\cite{Blum1988COLT,Goldman2000ICML,Ma2017icml} train learners alternately on distinct views with confident labels for unlabeled data. Coregularization style algorithms~\cite{Sindhwani2008icml,Ye2015cikm} regard disagreement between multiple views as a regularization term in the objective function. Multi-kernel learning algorithms~\cite{lanckriet2004learning,cortes2010two} exploit kernels that naturally correspond to different views and combine kernels to improve learning performance. In addition, multi-view graph clustering~\cite{kumar2011co} and multi-view subspace clustering~\cite{domeniconi2007locally} also belong to multi-view learning.
	
	Jointly learning a data classifier and a label aggregator can be seen as a coupled-view learning problem, but the above methods are mainly applied to unsupervised and semi-supervised learning, which cannot be directly applied to noisy supervised learning. This paper proposes trustable co-label learning in the learning-from-multiple-noisy-annotators setting, where co-labels act as the information exchange bridge between two classifiers of different views.
	
	\subsection{Hybrid Methods}
	Given the output of a set of base classifiers, rather than trying to find the best single learner, hybrid methods~(also called combination methods) aim to integrate learners to enhance the generalization ability.
	
	For nominal outputs, majority voting is the most common hybrid method~\cite{zhou2012ensemble}. Weighted voting~\cite{kim2011weight} assigns more weight to stronger classifiers for voting. The Naive Bayes combiner~\cite{titterington1981comparison} assumes that the classifiers are mutually independent given a true class label and takes the maximum label for the posterior probability. Other combination methods for such outputs include the BKS method~\cite{huang1993behavior}, Wernecke method~\cite{Wernecke1992bio}, SVD combination method~\cite{Merz99}, Bayesian classifier combination~\cite{KimG12}, etc.
	For numeric outputs, combiners can be classified into two main categories: nontrainable and trainable methods. Nontrainable combiner~\cite{K2004} has no extra parameters that need to be trained, including simple averaging, minimum/maximum/median rule, product rule, and generalized mean~\cite{DuboisP85}.
	Trainable combiners include weighted averaging~\cite{Hashem1994icnn,WangZZ09,Moral2015jes}, fuzzy integral~\cite{ChoK95}, stacking~\cite{wolpert1992stacked}, decision template method~\cite{kuncheva2001decision}, selection methods~\cite{Przybyla-Kasperek16}, etc. {Note that integrating classifiers can be regarded as a special form of the label aggregation problem, and some crowdsourcing methods can also be used for it.}
	
	Generally, there is no unique best combiner for all problems.
	In this paper, we are faced with two hybrid problems, i.e., how to aggregate multiple labels and how to combine predictions from two classifiers into co-labels. For the first problem, our approach learns a label aggregator modeled by a Naive Bayes classifier or multilayer neural networks. For the second one, we combine the label probabilities from both classifiers into a joint probability under the guidance of a small, trusted dataset.
	
	\section{Our Approach}
	\subsection{Problem Formulation}
	\myPara{Preliminaries.}
	 We are given a massive untrusted training dataset $\widetilde{\mathcal{D}}$ of $n$ instances drawn from $p(\bm{X}, \tilde{\bm{Y}})$. It includes data features $\bm{\mr{x}}=\{\bm{x}_i\}_{i=1}^{n}$ and multiple noisy labels $\bm{\tilde{\mr{y}}}=\{\tilde{\bm{y}}_i\}_{i=1}^{n}$, where $\tilde{\bm{y}}_i=(\tilde{y}^{(1)}_{i},\tilde{y}^{(2)}_{i},...,\tilde{y}^{(m)}_{i})$ is an $m$-dimensional vector from fixed $m$ labeling sources (it may have missing labels). $C$ is the size of label space, and $\tilde{y}^{(m)} \in \{ 1,2,...,C\}$.
	 
	We are given a small, trusted training dataset $\mathcal{D}$ of $u$ examples drawn from $p(\bm{X}, Y)$, which includes data features and corresponding clean labels.
	
	\myPara{Overview.} Following coupled-view learning, our approach needs to model two classifiers: 1) label aggregator $\phi_l \left(\tilde{\bm{y}}; \bm{\mr{w}}_l \right)$, which combines an $m$-dimensional weak label vector $\tilde{\bm{y}}$ from fixed $m$ annotators to produce an estimated distribution, and 2) data classifier $\phi_d \left(\bm{x}; \bm{\mr{w}}_d \right)$, which is a deep classifier that takes an instance feature $\bm{x}$ as the input and outputs a predicted distribution. Here, $\bm{\mr{w}}_d$ and $\bm{\mr{w}}_l$ are the learned parameters.
	
	To achieve effective and robust learning, our approach alternately conducts a classifier learning phase and prediction combination phase under the guidance of a small, trusted dataset~(see Fig. \ref{fig_first_case}), and the performance of the two training phases is progressive. {After that, it performs retraining on untrusted data with fixed co-labels and trusted data with given clean labels to attain the final data classifier.} Next, we introduce the two iterative phases in detail.
	
	\begin{figure*}[!t]
		\centering
		\subfloat[General Case]{\includegraphics[width=3.6in]{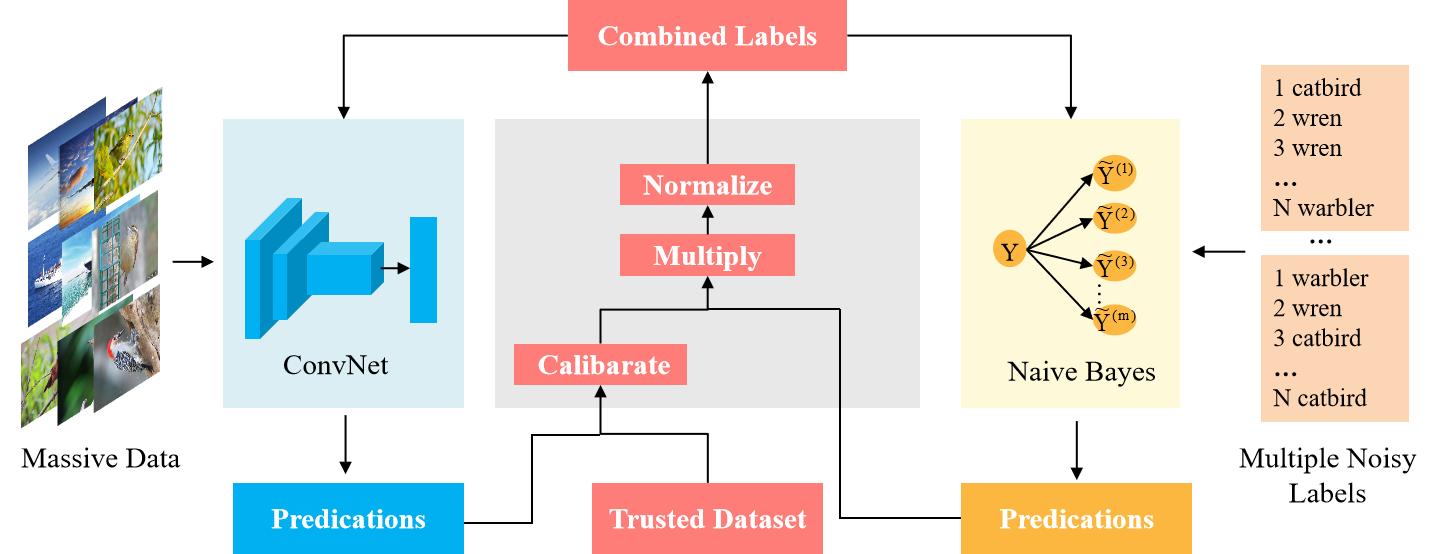}\label{fig_first_case}}
		\subfloat[A Special Complete Data Case]{\includegraphics[width=3.6in]{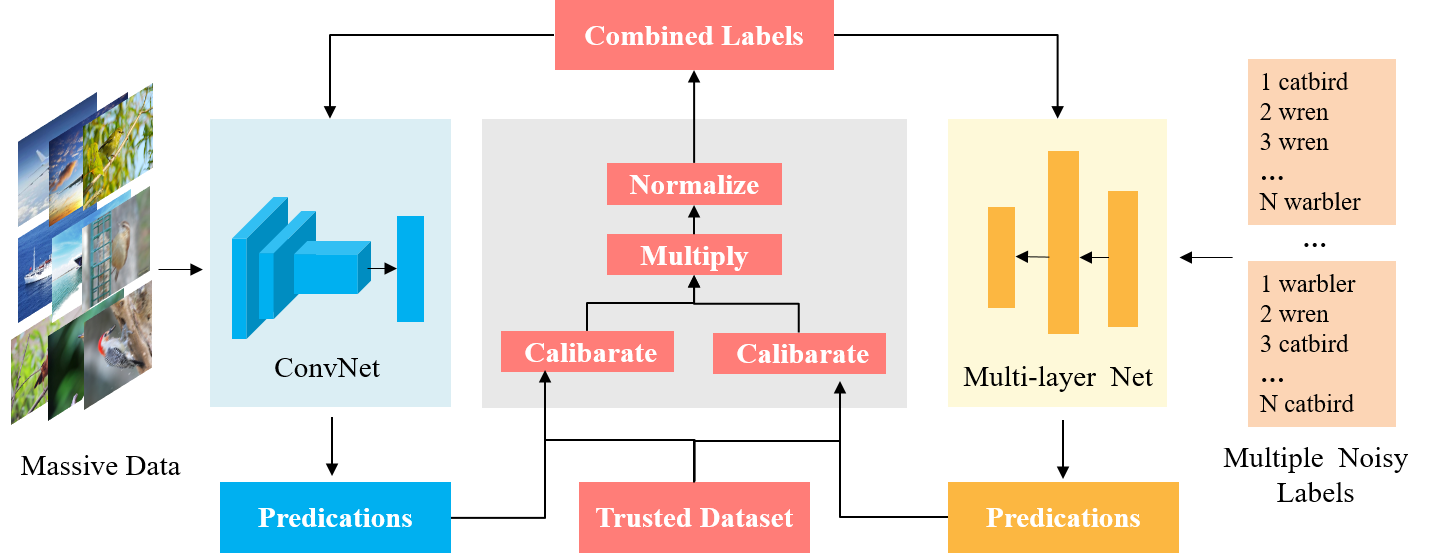}\label{fig_second_case}}
		\caption{The Framework of Trustable Co-label Learning from Multiple Noisy Annotators. (a) General case, where each annotator can only label a part of instances and the label aggregator is represented by a Naive Bayes classifier. (b) A special complete data case, where each annotator labels all instances and the label aggregator is modeled by multilayer neural networks.}
		\label{fig_sim}
	\end{figure*}

	\subsection{Classifier Learning}

	With the help of co-labels $\bm{\mr{y}}^c=\{\bm{y}^c_i\}_{i=1}^{n}$ that are combined from the predictions of $\phi_d$ and $\phi_l$ during training, we treat this weakly supervised learning problem as a supervised learning problem in the classifier learning phase:
	\begin{equation}
	\min_{\bm{\mr{w}}_d,\bm{\mr{w}}_l} \ell_d \left( \bm{\mr{x}}, \bm{\mr{y}}^c; \bm{\mr{w}}_d\right)+\ell_l \left( \bm{\tilde{\mr{y}}},\bm{\mr{y}}^c; \bm{\mr{w}}_l \right),
	\label{eq:cls_loss}
	\end{equation}
	where $\ell_d(.)$ and $\ell_l(.)$ are the loss functions for training $\phi_d$ and $\phi_l$, respectively.
	
	\myPara{Data classifier.}
	Due to its high capacity to learn from data, we regard a deep network $\phi_d \left(\bm{x}; \bm{\mr{w}}_d \right)$ as the data classifier. With the supervision of soft co-labels, its loss function can be written as Eq.~\eqref{eq:data classifier}:
	\begin{equation}
	\ell_d \left( \bm{\mr{x}}, \bm{\mr{y}}^c; \bm{\mr{w}}_d\right)= \sum_{i=1}^n \ell_{CE} \left( \phi_d (\bm{x}_i; \bm{\mr{w}}_d),\bm{y}^c_i\right),
	\label{eq:data classifier}
	\end{equation}
	where $\ell_{CE}(.)$ is the cross-entropy function and $\bm{y}^c_i$ is the $C$-dimensional co-label vector of instance $i$.
	
	\myPara{Label aggregator.}
	The label aggregator aims to infer an estimated label distribution for each instance from $m$ labeling sources. Similar to Li et al.~\cite{Li2020aaai}, with the assumption that every annotator is conditionally independent given true labels, we adopt a \emph{Naive Bayes} classifier to model label aggregation. When the confusion matrices $\bm{\pi}$ and the class prior probability $\bm{q}$ are known, ${\phi_l(\tilde{\bm{y}}_i;\bm{\mr{w}}_l= (\bm{\pi},\bm{q}))}_k=P(Y=k | \tilde{Y}=\tilde{\bm{y}}_i; \bm{\pi},\bm{q})$, the posterior probability of the true label of instance $i$ for class $k$ can be calculated as Eq. \eqref{eq:label aggregator 1}:
	\begin{equation}\label{eq:label aggregator 1}
	{\phi_l(\tilde{\bm{y}}_i;\bm{\pi},\bm{q})}_k=\frac{{q}_{k} \prod_{j=1}^{m}(\sum_{s=1}^{C} \mathbb{I}[\tilde{y}^{(j)}_{i}=s] {\pi}_{ks}^{(j)})}{\sum_{k^{\prime}=1}^{C}({q}_{k^{\prime}} \prod_{j=1}^{m}(\sum_{s=1}^{C} \mathbb{I}[\tilde{y}^{(j)}_{i}=s] {\pi}_{k^{\prime} s}^{(j)}))},
	\end{equation}
	where $\tilde{y}^{(j)}_{i}$ is the noisy label of instance $i$ from annotator $j$; $\mb{I}[.]$ is the indicator function, which takes 1 if the event is true and 0 otherwise; ${\pi}_{ks}^{(j)}$ is the probability of misclassifying class $k$ into class $s$ for annotator $j$; and $q_k$ is the prior probability of class $k$.
	
	With the supervision of soft co-labels, its loss function can be written as:
	\begin{equation}\label{eq:label aggregator 2}
	\ell_l \left( \bm{\tilde{\mr{y}}},\bm{\mr{y}}^c; \bm{\pi},\bm{q} \right) = \sum_{i=1}^n \ell_{NLL} \left( \phi_l(\tilde{\bm{y}}_i;\bm{\pi},\bm{q}),\bm{y}^c_i\right),
	\end{equation}
	where $\ell_{NLL}(.)$ is a negative log-likelihood function and $\bm{y}^c_i$ is the $C$-dimensional co-label vector of instance $i$.
	
	To solve this problem, $\bm{\pi}$ is acquired by Eq. \eqref{eq:label aggregator 3}, and $\bm{q}$ is estimated from trusted data $\mathcal{D}$:
	\begin{equation}\label{eq:label aggregator 3}
	{\pi}_{k s}^{(j)}=\frac{\sum_{i=1}^n  \mathbb{I}[\tilde{y}^{(j)}_{i}=s] y^c_{ik}}{\sum_{i=1}^n y^c_{ik}},
	\end{equation}
	where $y^c_{ik}$ is the co-label for class $k$ of instance $i$.
	
	\subsection{Prediction Combination}
	
	As the simple intuition illustrated in Fig. \ref{fig:motivation},
	we want to make use of the trusted data to combine the predictions $\phi_d(\bm{x}_i)$ and $\phi_l(\tilde{\bm{y}}_i)$ into more trustable co-labels $\bm{y}^c_i$ in each training iteration to make the co-label learning efficient and robust. This practice agrees with the observations in recent works~\cite{ZhuL021,XiaL00WGC21} that the prediction from deep networks during training contains information about the true label distribution, which can be used to reduce label noise through label correction~\cite{Khetan2018iclr,tanaka2018joint} or sample selection~\cite{Han2018NIPS, Jiang2018ICML,bai2020me}.

	Our prediction combination process is based on the perspective of probability. In classical machine learning, soft prediction from classifiers can be seen as an estimation of the posterior probability. However, as modern neural networks are usually poorly calibrated~\cite{GuoPSW17}, $\phi_d(\bm{x}_i)$, the prediction from deep networks, cannot be effectively representative of the probability, which makes the confidences from the data classifier and label aggregator hard to compare and combine directly. {This is where a small, trusted dataset helps, i.e., we use it to calibrate the prediction from deep neural networks for combination.} In the experiments, we use the isotonic regression method~\cite{ZadroznyE02} to perform calibration. Fig.~\ref{first_calibration} clearly shows the important effect of calibration. As reported, the VGG-16 model learned with initialized co-labels on the IND-3 dataset is very poorly calibrated, which means that its prediction confidence cannot reflect the ground truth correctness likelihood; in contrast, after calibration, the Expected Calibration Error~(ECE) of the data classifier decreases from 9.87 to 3.67, and its accuracy even increases from 30.09\% to 43.29\%. {Subsequent experiments demonstrate that such calibration via trusted data plays a key role in guiding trustable prediction combination and further encourages robust classifier learning.}

	\begin{figure}[t]
	\centering
	\subfloat[]{\includegraphics[width=1.7in]{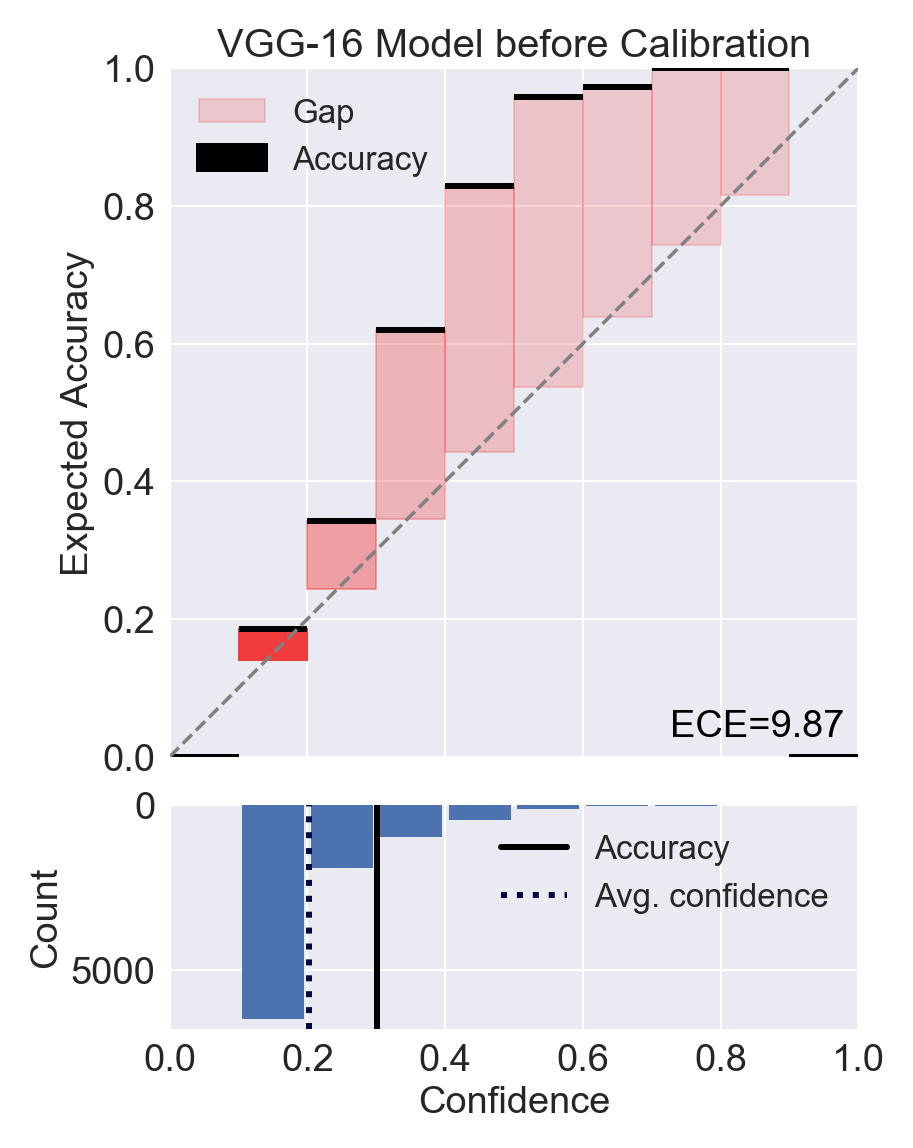}}
	\subfloat[]{\includegraphics[width=1.7in]{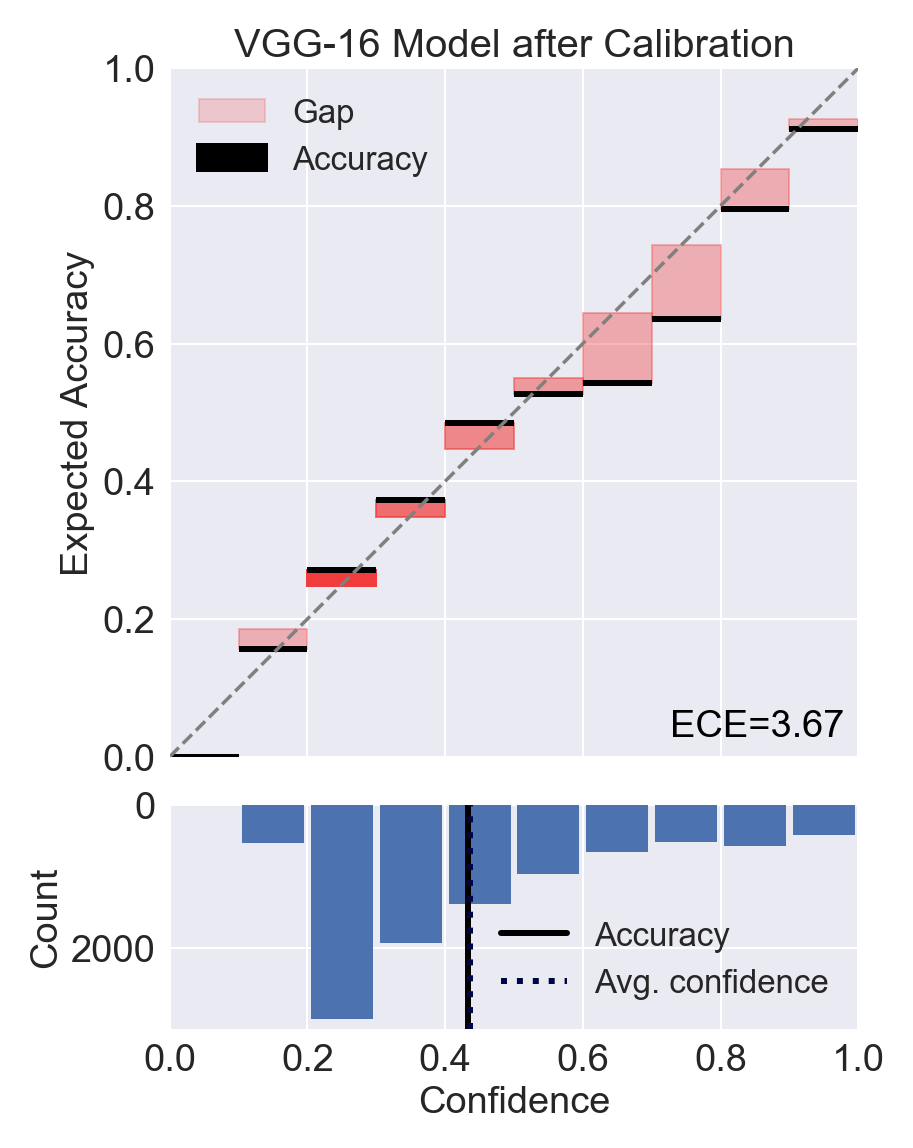}}
	\caption{Reliability diagrams (top) and confidence histograms (bottom) for the VGG-16 data classifier before calibration~(a) and after calibration~(b) in the first prediction combination phase on the IND-3 dataset.}
	\label{first_calibration}
    \end{figure}
	
	Therefore, with trusted data $\mathcal{D}$, we calibrate $\phi_d(\bm{x}_i)$ to let it estimate the true label likelihood, which means we assume the well-calibrated prediction as $\hat{P}(Y|\bm{S}=\bm{s}_i,\mathcal{D})$, where $\bm{s}_i=\phi_d(\bm{x}_i)$.
In addition, $\phi_l(\tilde{\bm{y}}_i)$ the prediction from Naive Bayes classifier can be seen as the estimated distribution $\hat{P}(Y| \tilde{\bm{Y}}=\tilde{\bm{y}}_i)$.
	
	With the well-estimated distributions, we can combine two predictions into a joint distribution based on the conditionally independent assumption, which means that assuming $\tilde{\bm{Y}}$ and $\{\bm{S},\mathcal{D}\} $ are independent given true label $Y$,  $P(Y|\bm{S},\mathcal{D},\tilde{\bm{Y}})$ can be inferred by Eq. \eqref{eq:metric1}:
	\begin{equation}
	\begin{aligned} P(Y| \bm{S},\mathcal{D},\tilde{\bm{Y}}) &=\! \frac{P(Y) p(\bm{S},\mathcal{D},\tilde{\bm{Y}}|Y)}{p(\bm{S},\mathcal{D},\tilde{\bm{Y}})} \\ &=\! \frac{P(Y)}{p(\bm{S},\mathcal{D},\tilde{\bm{Y}})} p(\bm{S},\mathcal{D}|Y) P(\tilde{\bm{Y}}|Y) \\ &=\! \frac{P(Y)}{p(\bm{S},\mathcal{D},\tilde{\bm{Y}})}\frac{P(Y|\bm{S},\mathcal{D})p(\bm{S},\mathcal{D})}{P(Y)} \frac{P(Y|\tilde{\bm{Y}})P(\tilde{\bm{Y}})}{P(Y)} \\ &=\!\frac{p(\bm{S},\mathcal{D})P(\tilde{\bm{Y}})}{p(\bm{S},\mathcal{D},\tilde{\bm{Y}})}\frac{P(Y|\bm{S},\mathcal{D})P(Y|\tilde{\bm{Y}})}{P(Y)}.
	\end{aligned}
	\label{eq:metric1}
	\end{equation}
	As $P(Y| \bm{S},\mathcal{D},\tilde{\bm{Y}}) \propto {P(Y|\mathbf{S},\mathcal{D})P(Y|\tilde{\mathbf{Y}})}/{P(Y)}$, we acquire the combined distribution by Eq. \eqref{eq:metric2}: 
	\begin{equation}\hat{P}(Y|\bm{s}_i,\mathcal{D},\tilde{\bm{y}}_i)=\operatorname{ Nor }\left[\left(\frac{\hat{P}(Y=k|\bm{s}_i,\mathcal{D}) \cdot \hat{P}(Y=k|\tilde{\bm{y}}_i)}{q_k}\right)_{k}\right],
	\label{eq:metric2}
	\end{equation}
	where Nor$[\mathbf{r}]:=\frac{\mathbf{r}}{\sum_{k} r_{k}}$.

After that, in each prediction combination phase, the co-labels are updated by the combined distributions. Note that although the independent assumption is not satisfied in real situations, the soft distributions are empirically verified to be more reliable than classifiers in our experiments.
	
	\subsection{Alternate Optimization}
	The whole approach is shown in Alg.~\ref{alg:algorithm1}. After the co-labels are initialized by majority voting, alternate optimization of the two training phases is conducted. During training iteration, reliable combined distributions make classifier learning more robust, and then better classifiers encourage more reliable combination in the next iteration. By such alternate optimization, our approach progressively attains more trustable co-labels. In addition, as the deep classifier may overfit noise when the co-labels are highly noisy in the first few rounds, to mitigate this impact and further utilize trusted data, after the alternate optimization stage, our approach reinitializes the data classifier and performs the retraining stage, which learns deep networks on untrusted data with fixed co-labels and trusted data with given clean labels. {In this way, the small, trusted dataset not only provides the original clean target for learning but also serves as a guide to alleviate the unreliability of co-labels.}
	
	\begin{algorithm}[t]
		\small
		\renewcommand{\algorithmicrequire}{\textbf{Input:}}
		\renewcommand{\algorithmicensure}{\textbf{Output:}}
		\caption{Trustable Co-label Learning from Multiple Noisy Annotators}
		\label{alg:algorithm1}
		\begin{algorithmic}[1]
			\REQUIRE untrusted dataset $\widetilde{\mathcal{D}}=\{\bm{\mr{x}}, \bm{\tilde{\mr{y}}}\}$, trusted dataset $\mathcal{D}$; max iteration round $T$.
			\ENSURE learned parameters $\bm{\mr{w}}_{d}$ of deep classifier $\phi_{d}$.
			\STATE Initialize co-labels $\bm{{y}}^c$ by majority voting from $\bm{\tilde{\mr{y}}}$ and initialize the parameters $\bm{\mr{w}}_{d}$ of $\phi_{d}$.
			\FOR {$t=1,2,...,T$}
			\STATE Train $\phi_l$ on $\{\bm{\tilde{\mr{y}}},\bm{{y}}^c\}$.

			\STATE Combine predictions and update $\bm{{y}}^c$ by Eq. \eqref{eq:metric2}.
			\STATE Train $\phi_d$ on $\{\bm{\mr{x}},\bm{{y}}^c\}$.
			\STATE Combine predictions and update $\bm{{y}}^c$ by Eq. \eqref{eq:metric2}.
			\ENDFOR
			\STATE Reinitialize $\phi_{d}$ and retrain it on $\{\bm{\mr{x}}, \bm{{y}}^c\}$ and $\mathcal{D}$ .
			\STATE \textbf{return} $\bm{\mr{w}}_{d}$.
		\end{algorithmic}
	\end{algorithm}

\subsection{Our Approach for A Special Complete Data Case}
	Typically, when instances are labeled by multiple non-expert human workers, the label data $\bm{\tilde{\mr{y}}}$ are very sparse, which means that each worker annotates only a part of the samples, leading to many missing labels. This property limits the modeling method of label aggregation since some powerful modeling methods, such as multilayer neural networks, cannot handle the serious missing problem well.
	However, recently, with the huge need for large-scale labeled datasets, the deep learning community has a great interest in making use of automatic labeling methods, such as weak classifiers from small-scale data~\cite{Li2017iccv,9113752} or similar domains~\cite{RahmanKB20}, messy user tags from social media~\cite{9337209,YadatiLLH18}, matched results from search engines~\cite{DengJTGL14, LiSLZ18}, relations from knowledge bases~\cite{ChaudharyGPC20} and other human-free labeling sources. The good news is that it is easier to label all instances without missing by those automatic labeling sources in some scenarios (e.g., social network tagging~\cite{ChaC12,PangN15} and financial analysis~\cite{ait2010high}).
	This motivates us to improve our approach (see Fig. \ref{fig_second_case}) by modeling the label aggregator as a multilayer neural network for a special complete data case where every annotator labels all training data. For simplicity, we introduce the differences of the improved approach (denoted as TCLS) from the original TCL as follows:
	
	\myPara{Training dataset.} We are given a massive untrusted dataset $\widetilde{\mathcal{D}}=\{\bm{\mr{x}}, \bm{\tilde{\mr{y}}}\}$, where $\bm{\tilde{\mr{y}}}$ is complete (i.e., it has no missing labels), and a small trusted dataset $\mathcal{D}$ drawn from $p(\bm{X}, \tilde{\bm{Y}}, Y)$, which includes data features, complete multiple noisy labels, and clean labels.
	
	\myPara{Label aggregator learning.} To enhance modeling ability, the label aggregator $\phi_l$ is represented by multilayer neural networks. Its loss function can be written as:
	\begin{equation}\label{eq:label aggregator 4}
	\ell_l \left( \bm{\tilde{\mr{y}}},\bm{\mr{y}}^c; \bm{\mr{w}}_{l} \right) = \sum_{i=1}^n \ell_{CE} \left( \phi_l(\tilde{\bm{y}}_i;\bm{\mr{w}}_{l}),\bm{y}^c_i\right).
	\end{equation}

\begin{algorithm}[t]
	\small
	\renewcommand{\algorithmicrequire}{\textbf{Input:}}
	\renewcommand{\algorithmicensure}{\textbf{Output:}}
	\caption{Trustable Co-label Learning for a Special Complete Data Case}
	\label{alg:algorithm2}
	\begin{algorithmic}[1]
		\REQUIRE untrusted dataset $\widetilde{\mathcal{D}}=\{\bm{\mr{x}}, \bm{\tilde{\mr{y}}}\}$, trusted dataset $\mathcal{D}$; max iteration round $T$.
		\ENSURE learned parameters $\bm{\mr{w}}_{d}$ of deep classifier $\phi_{d}$.
		\STATE Initialize co-labels $\bm{{y}}^c$ from the predictions of Naive Bayes classifier, which is trained on trusted data $\mathcal{D}$, and initialize the parameters $\bm{\mr{w}}_{d}$ of $\phi_{d}$ and $\bm{\mr{w}}_{l}$ of $\phi_{l}$.
		\FOR {$t=1,2,...,T$}
		\STATE Train $\phi_l$ on $\{\bm{\tilde{\mr{y}}},\bm{{y}}^c\}$.
		\STATE Combine predictions and update $\bm{{y}}^c$ by Eq. \eqref{eq:metric3}.
		\STATE Train $\phi_d$ on $\{\bm{\mr{x}},\bm{{y}}^c\}$.
		\STATE Combine predictions and update $\bm{{y}}^c$ by Eq. \eqref{eq:metric3}.
		\ENDFOR
		\STATE Reinitialize $\phi_{d}$ and retrain it on $\{\bm{\mr{x}}, \bm{{y}}^c\}$ and $\mathcal{D}$ .
		\STATE \textbf{return} $\bm{\mr{w}}_{d}$.
	\end{algorithmic}
\end{algorithm}
	
	\myPara{Prediction combination.}
	{Following the same inspiration as TCL, we reannotate co-labels by effectively utilizing trusted data. First, we calibrated $\phi_d(\bm{x}_i)$ and $\phi_l(\tilde{\bm{y}}_i)$ via the trusted dataset $\mathcal{D}$,} and then regard them as $\hat{P}(Y|\bm{S}=\bm{s}_i,\mathcal{D})$ and $\hat{P}(Y|\bm{V}=\bm{v}_i,\mathcal{D})$, respectively, where $\bm{s}_i=\phi_d(\bm{x}_i)$ and $\bm{v}_i=\phi_l(\tilde{\bm{y}}_i)$.
	Assuming $\bm{V}$ and $\bm{S} $ are conditionally independent given $Y$, $p(\bm{S},\mathcal{D}|Y)\approx p(\mathcal{D})p(\bm{S}|Y)$ and $p(\bm{V},\mathcal{D}|Y)\approx p(\mathcal{D})p(\bm{V}|Y)$, we can obtain $P(Y| \bm{S},\mathcal{D},\bm{V}) \propto {P(Y|\mathbf{S},\mathcal{D})P(Y|\mathbf{V},\mathcal{D})}/{P(Y)}$; thus, we update co-labels by Eq. \eqref{eq:metric3}:
	\begin{equation}\hat{P}(Y|\bm{s}_i,\mathcal{D},\bm{v}_i)=\operatorname{ Nor }\left[\left(\frac{\hat{P}(Y=k|\bm{s}_i,\mathcal{D}) \cdot \hat{P}(Y=k|\bm{v}_i,\mathcal{D})}{q_k}\right)_{k}\right].
	\label{eq:metric3}
	\end{equation}
	In addition, to utilize the existing knowledge, we initialize co-labels from the predictions of Naive Bayes classifier, which is trained on $\mathcal{D}$.

	With the high modeling capacity of deep networks, it is natural that the performance of label aggregator benefits from it, and we verify this in our experiments. The algorithm process is shown in Alg.~\ref{alg:algorithm2}.
	
	\section{Experiments}
	
\begin{table*}[t]
	\small
	\caption{The generated datasets by simulating different independent annotators on CIFAR10.}

	\centering

	\begin{tabular}{cccc}

		\hline Datasets & 1st group of annotators & 2nd group of annotators & 3rd group of annotators \\

		\hline

		IND-1 & Symmetry, $\varepsilon=0.8$ & Symmetry, $\varepsilon=0.7$ & Pair, $\varepsilon=0.45$ \\

		IND-2 & Symmetry, $\varepsilon=0.85$ & Pair, $\varepsilon=0.45$ & Classwise, correct class 1 \\

		IND-3 & Symmetry, $\varepsilon=0.8$ & Symmetry, $\varepsilon=0.7$ & Classwise, correct class 7,8,9 \\

		IND-4 & Symmetry, $\varepsilon=0.6$ & Symmetry, $\varepsilon=0.7$ & Classwise, correct class 3,5,7 \\

		\hline

	\end{tabular}

	\label{groups}

\end{table*}

	To verify the effectiveness and robustness of our approach, we conduct experiments on both synthetic and real datasets under two settings: learning from non-expert humans and learning from auto-labeling sources.
	
	\subsection{Learning from Non-expert Humans}
	Learning from non-expert humans obtains data labeled by more than one person. It is a typical and well-known setting for learning from multiple noisy annotators, and the label data are usually sparse. In this section, we conduct experiments under such setting on one synthetic dataset (CIFAR10~\cite{Krizhevsky09tr}) and two real datasets (LabelMe-AMT~\cite{Rodrigues2018aaai} and CUBShape~\cite{WelinderEtal2010}) to evaluate our TCL.
	
	\myPara{Datasets.}~\textbf{CIFAR10} is a 10-class image classification dataset that consists of 50K training images and 10K validation images. We retain 1K samples (each class has 100 samples) of the training data for trusted data and corrupt the other data manually by the confusion matrix $Q$, where $Q_{ij}=P(\tilde{Y}=j|Y=i)$, given that one noisy label $\tilde{Y}$ is flipped from clean label $Y$. We produce noisy labels with three kinds of confusion matrices, including 1) \textbf{symmetry flipping}, which simulates that the annotator may choose false labels uniformly at random with probability $\varepsilon$; 2) \textbf{pair flipping}, which imitates the annotator who may confuse similar classes with probability $\varepsilon$; and 3) \textbf{class-wise flipping}, which simulates that the annotator only does good labeling in particular classes but chooses labels uniformly at random for other classes.
	In our problem setting, the untrusted dataset is labeled by 30 noisy annotators, and each instance has 3 weak labels from 
three randomly chosen different annotators. All annotators are produced from 3 different confusion matrices $Q$, i.e., one matrix produces 10 annotators. To cover more cases, we design 4 sets of independent annotators in such a setting, resulting in 4 noisy datasets (see Tab. \ref{groups}). For training, we pad the pictures on all sides by 4 pixels, randomly crop them by the size of $32 \times 32$, apply random horizontal flip, and finally normalize them. For validation, we only normalize the pictures.
	
	\textbf{LabelMe-AMT} is a real-world 8-class image classification dataset. It consists of a total of 2,688 images, where 1,000 of them are used to obtain noisy labels by an average of 2.5 workers per image (59 workers in total) from Amazon Mechanical Turk. 80 images (each class 10 samples) are used for trusted data, while 1608 images are used for validation. We follow the image preprocessing method in~\cite{Rodrigues2018aaai}.
	
	The real dataset \textbf{CUBShape} is adapted from the shape task (the shape is perching-like or not) of the CUB-200-2010 dataset~\cite{WelinderEtal2010}. It contains the binary labeling task to label the shape for 6,033 bird images from Amazon Mechanical Turk. There are approximately 500 users contributing labels, and each image receives 5 labels. We collect ground truth from whatbird.com for evaluation. We retain 5000 images for untrusted data, 100 images for trusted data, and 933 images for validation data. For training, we apply random horizontal flip, cropping~($448 \times 448$) and cutout and finally normalize them. For validation, we only apply center cropping and normalization.
	
	\myPara{Implementation.}
	For synthetic CIFAR10 datasets, we adopt the VGG-16 networks~\cite{SimonyanZ14a} to model the data classifier. Our TCL approach trains two classifiers for $T=60$ iterations and then retrains the data classifier for 60 epochs.
	In each iteration, the data classifier is trained for 1 epoch. During the alternate optimization stage and retraining stage, we use SGD with a batch size of 128, a momentum of 0.9, a weight decay of 0.0005, and an initial learning rate of 0.1. The learning rate is divided by 10 after 40 epochs and 50 epochs (for a total of 60 epochs).
	
	For the real LabelMe-AMT dataset, we use the pretrained CNN layers of the VGG-16 network and apply only one FC layer (with 128 units and ReLU activations) and one output layer on top with 50$\%$ dropout. Our TCL approach trains two classifiers for $T=29$ iterations, and retrains the data classifier for 30 epochs. In the first iteration round, the data classifier is trained for 2 epochs, and in the other round for 1 epoch (30 epochs in total). We used the Adam optimizer with a batch size of 128, a learning rate of 0.0001, and betas of (0.9, 0.999).
	
	For real the CUBShape dataset, we also use the pretrained CNN layers of the VGG-16 network and apply only one FC layer (with 128 units and ReLU activations) and one output layer on top with 50$\%$ dropout. Our TCL approach trains two classifiers for $T=29$ iterations, and then retrains the data classifier for 10 epochs. In the first iteration round, the data classifier is trained for 2 epochs, and in the other round, it is trained for 1 epoch (30 epochs in total). During the alternate optimization stage and retraining stage, we use SGD with a batch size of 16, a momentum of 0.9, a weight decay of 0.0005, and an initial learning rate of 0.01. The learning rate is divided by 10 after 20 epochs.

	\begin{table}[t]
	\small
	\caption{Validation accuracy ($\%$) on synthetic CIFAR10 datasets with independent annotators. "F" denotes fine-tuning on trusted data. The minimal improvement is also given.}
	\centering

	\begin{tabular}{p{2.4cm}<{\centering}p{1.1cm}<{\centering}p{1.1cm}<{\centering}p{1.1cm}<{\centering}p{1.1cm}<{\centering}}

		\hline Approach & IND-1  & IND-2 & IND-3 & IND-4 \\

		\hline 	DL-MV &  71.77 & 52.19 & 44.72 & 68.94  \\

		{DL-DS} &  {89.88} & {80.89} & {62.03} & {83.39}  \\

		{DL-IBCC} &  {90.43} & {77.05} & {50.42} & {84.08}  \\

		{DL-CRH} &  {86.80} & {88.23} & {37.68} & {76.82}  \\

		AggNet &  91.42 & 88.79 & 82.23 & 88.81 \\	 	

		CrowdLayer &  89.32 & 88.42 & 80.27 & 83.96 \\

		MBEM &  90.90 & 89.77 & 79.63 & 87.64   \\

		Max-MIG &  90.29 & 88.59 & 83.47 & 88.00 \\

		CVL&  88.72 & 86.93 & 76.12 & 84.37 \\

		\hline DL-MV+F &  81.68 & 72.68 & 61.06 & 80.02   \\

		{DL-DS+F} &  {89.93} & {82.72} & {64.03} & {83.82}  \\

		{DL-IBCC+F} &  {90.46} & {82.40} & {64.97} & {84.78}  \\

		{DL-CRH+F} &  {90.36} & {90.43} & {39.81} & {77.12}  \\

		AggNet+F &  \underline{91.56} & 89.57& 82.33 & \underline{88.90} \\	 	

		CrowdLayer+F &  89.60 & 90.64 & 80.57 & 83.99 \\

		MBEM+F &  90.96 & \underline{90.66} & 80.08 & 87.84 \\

		Max-MIG+F &  90.49 & 90.29 & \underline{83.58} & 88.12 \\

		CVL+F &  89.63 & 89.32 & 79.31 & 84.45 \\

		\hline 	Our TCL &  \textbf{92.50} & \textbf{92.86} & \textbf{86.78} & \textbf{91.16}  \\

		\hline 	        Min$\uparrow$  & 0.94 &  2.20&  3.20&2.26  \\

		\hline

	\end{tabular}

	\label{cifar10_ind}
\end{table}	

	\myPara{Results.}
	{First, we compare our TCL approach with several label aggregation methods without using data features to train the deep neural network, whose estimated probabilities are used directly as learning targets (as done in knowledge distillation~\cite{GouYMT21}), } including (i) \textbf{DL-MV} that trains a DNN on the result of (hard) majority voting~\cite{IpeirotisPSW14}; {(ii) \textbf{DL-DS} that trains a DNN on the result of the Dawid-Skene estimator
	~\cite{Dawid1979Maximum}; (iii) \textbf{DL-IBCC} that trains a DNN on the result of the independent Bayesian classifier combination~\cite{KimG12}~\footnote{{Since the original version of IBCC~\cite{KimG12} using Gibbs sampling is expensive to run and therefore unusable when
	thousands of labels are present, we use its variant using variational Bayes~\cite{SimpsonRPS13}.}}; and (iv) \textbf{DL-CRH} that trains a DNN on the result of CRH model~\cite{LiLGZFH14}.
	Second, we compare it with the approaches that combine
	data classifier learning and annotator modeling in a joint manner,}
	including (i) \textbf{AggNet}~\cite{Albarqouni2016TMI} that uses EM algorithm to jointly estimate workers' skills and a data classifier; (ii) \textbf{Crowd Layer}~\cite{Rodrigues2018aaai} that adds a crowd layer to the output of a common network to model confusion matrices; (ii) \textbf{MBEM}~\cite{Khetan2018iclr} is an improved EM algorithm that rewrites the EM likelihood and regards the estimated true labels as hard labels; (iv) \textbf{Max-MIG}~\cite{CaoXKW19} is an information theoretic method, which finds the information intersection between two classifiers; and (v) \textbf{CVL}~\cite{Li2020aaai} is a coupled-view method, which introduces several effective learning schemes to enhance robustness to label noise. In addition, since our method uses information from clean trusted data, for a fair comparison, we conduct additional fine-tuning on the trusted data based on these pretrained baselines. All the results are reported as the average figures of three trials. {In our approach, we use the isotonic regression method~\cite{ZadroznyE02} or its multi-class version~\cite{GuoPSW17} to perform calibration during training.}

	Tab. \ref{cifar10_ind} shows the results on four synthetic crowdsourcing CIFAR10 datasets, where some observations can be concluded.
	First, the common baseline, DL-MV, which directly learns with the aggregated labels by majority voting, performs poorly on all datasets. This implies that effective label aggregation is crucial to improving the performance of the data classifier.
	{Second, we can see that except for DL-MV, the three label aggregation methods are not stable across different datasets, and there is no label aggregation algorithm that outperforms others consistently, which is in accord with the observations in related work~\cite{ZhengLLSC17}.
	Third, the performance of label aggregation methods without using instance features is weaker than the methods that jointly learn the data classifier and annotator model.} Fourth, the simple fine-tuning technique can help, but it is not very effective since the amount of trusted data is small.
	Last, our TCL approach outperforms other benchmarks on all four datasets under the evaluation of validation accuracy, {showing TCL is more data-efficient than employing the fine-tuning technique in the existing methods.}

Fig. \ref{ind_3} clearly shows the different iterative learning processes of TCL without and with calibration on the IND-3 dataset. TCL without calibration does not combine prediction into more reliable co-labels, and classifiers eventually fit to more label noise. In contrast, with calibration via trusted data, our TCL approach achieves effective mutual improvement of classifier learning and prediction combination during training, i.e.,  the pure co-labels make models less overfit to false labels, and the co-labels become purer as the model improves; finally, both reliable co-labels and clean labels are used to retrain a network, which provides more improvement~(see Section \ref{ablation}). Hence, the great performance of our approach is due to the reliable prediction combination via the effective usage of trusted data.

	\begin{table}[t]
	\small

	\caption{Validation accuracy (\%) on LabelMe-AMT and CUBShape datasets. "F" denotes fine-tuning on trusted data.}

	\centering

	\begin{tabular}{p{2.4cm}<{\centering}p{2.4cm}<{\centering}p{2.4cm}<{\centering}}

		\hline Approach  & LabelMe-AMT &  CUBShape\\

		\hline DL-MV & 79.35 ($\pm$ 0.48)&  91.42 ($\pm$ 0.18)\\

		{DL-DS} & {82.48 ($\pm$ 0.13)}&  {92.00 ($\pm$ 0.16)}\\

		{DL-IBCC} & {81.92 ($\pm$ 0.07)}&  {92.07 ($\pm$ 0.11)}\\

		{DL-CRH} & {80.12 ($\pm$ 0.25)} &  {91.28 ($\pm$ 0.16)}\\

		AggNet  & 84.82 ($\pm$ 0.16)& 92.18 ($\pm$ 0.10)\\

		CrowdLayer & 81.42 ($\pm$ 3.34)& 91.92 ($\pm$ 0.06)\\

		MBEM & 79.63 ($\pm$ 3.74)&92.06 ($\pm$ 0.18)\\

		MAX-MIG & 85.60 ($\pm$ 0.14)&91.96 ($\pm$ 0.38)\\

		CVL & 86.04 ($\pm$ 0.34)&91.85 ($\pm$ 0.10)\\

		\hline DL-MV+F & 84.21 ($\pm$ 0.30)&91.85 ($\pm$ 0.28)\\

		{DL-DS+F} & {86.27 ($\pm$ 0.13)}&  {92.21 ($\pm$ 0.16)}\\

		{DL-IBCC+F} & {86.35 ($\pm$ 0.09)}&  {92.25 ($\pm$ 0.22)}\\

		{DL-CRH+F} & {85.32 ($\pm$ 0.11)}&  {91.46 ($\pm$ 0.16)}\\

		AggNet+F & 86.04 ($\pm$ 0.29)& 92.18 ($\pm$ 0.10)\\

		CrowdLayer+F & 85.38 ($\pm$ 1.40)&91.99 ($\pm$ 0.16)\\

		MBEM+F & 85.55 ($\pm$ 0.74)&92.28 ($\pm$ 0.18)\\

		MAX-MIG+F & 86.19 ($\pm$ 0.06)&91.99 ($\pm$ 0.32)\\

		CVL+F & 86.89 ($\pm$ 0.25)&92.06 ($\pm$ 0.10)\\

		\hline Our TCL & \textbf{88.09 ($\pm$ 0.12)}&\textbf{92.64 ($\pm$ 0.06)}\\


		\hline

	\end{tabular}

	\label{labelme_result}

\end{table}

	\begin{figure*}[t]
	\centering
	\subfloat[annotator  1]{\includegraphics[width=2.55in]{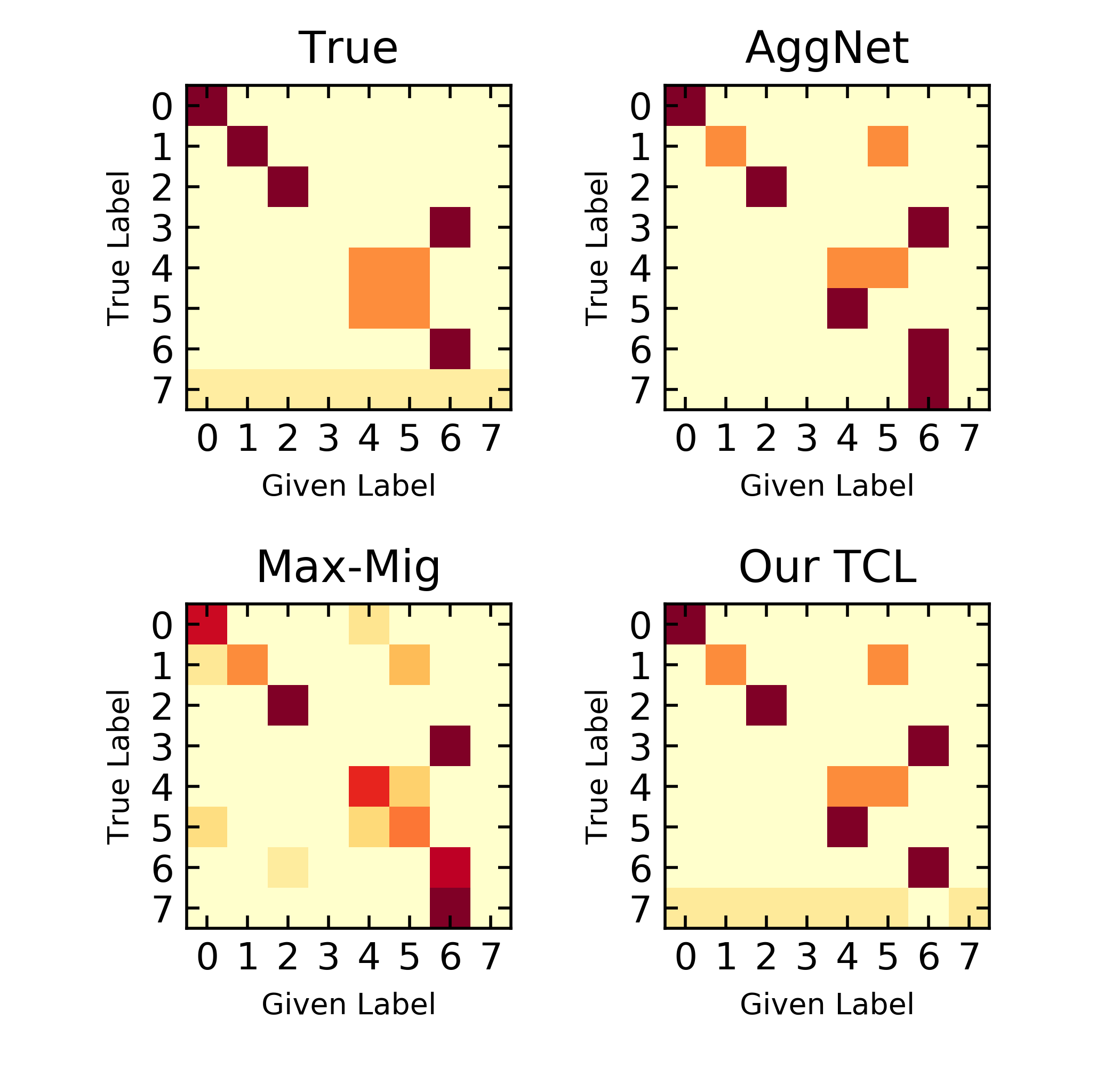}}
	\hspace{-8mm}
	\subfloat[annotator  20]{\includegraphics[width=2.55in]{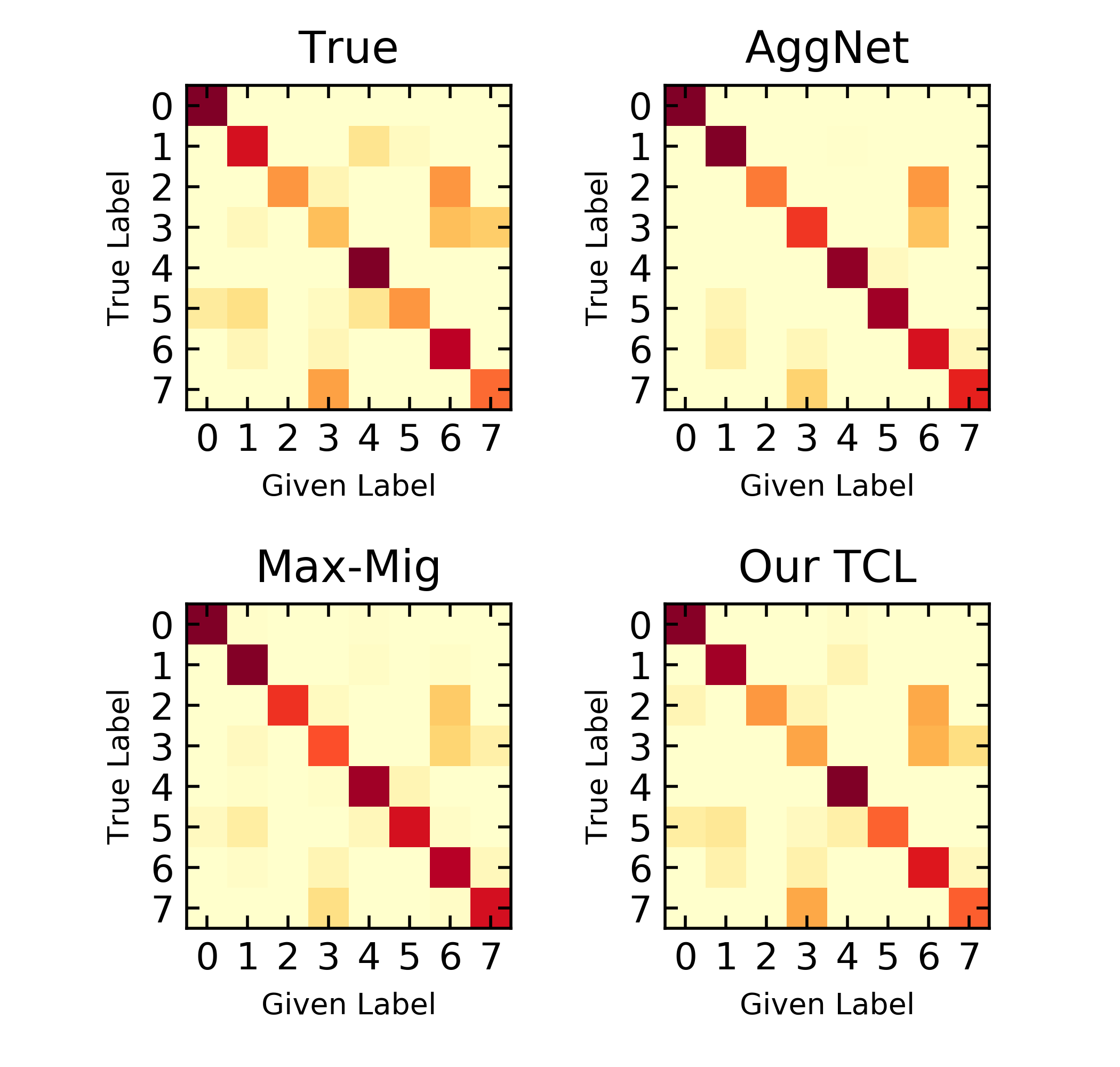}}
	\hspace{-8mm}
	\subfloat[annotator  39]{\includegraphics[width=2.55in]{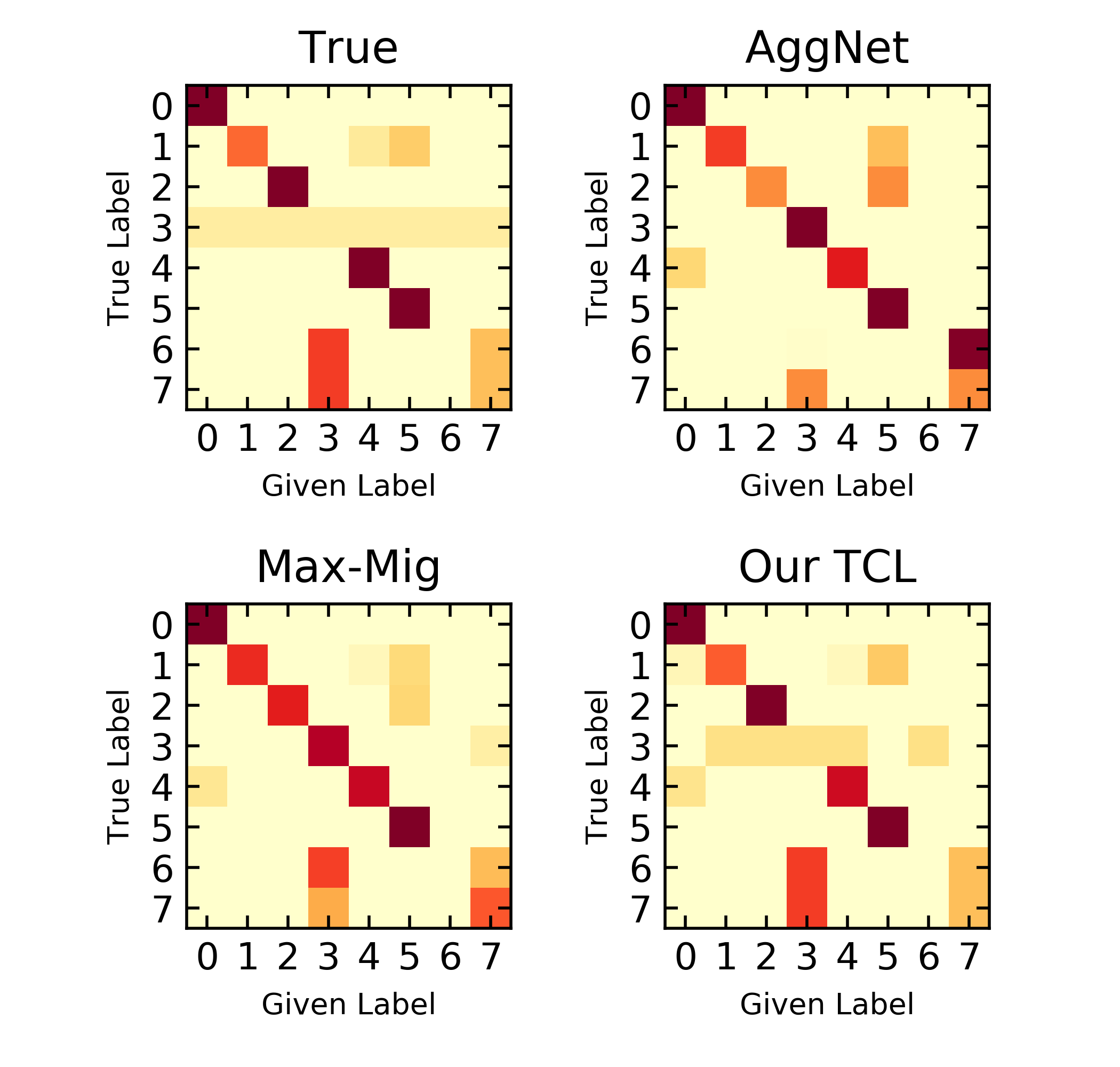}}
	\caption{Comparison between the true confusion matrices and the estimated confusion matrices of three annotators by different methods on the real-world dataset LabelMe-AMT.}
	\label{labelme_matrices}
\end{figure*}

\begin{figure}[t]
	\centering
	\subfloat[]{\includegraphics[width=1.7in]{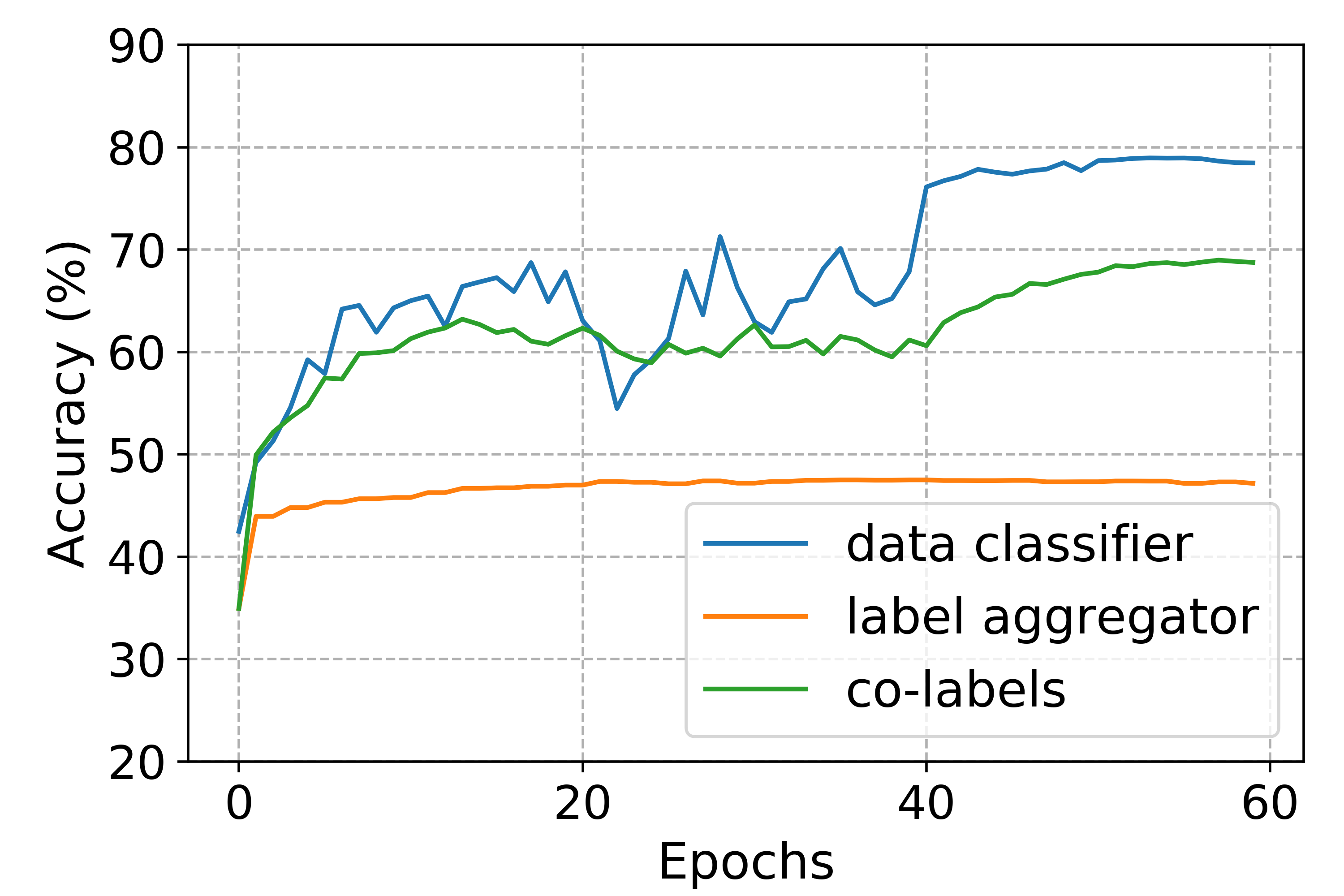}} 
	\hfil
	\subfloat[]{\includegraphics[width=1.7in]{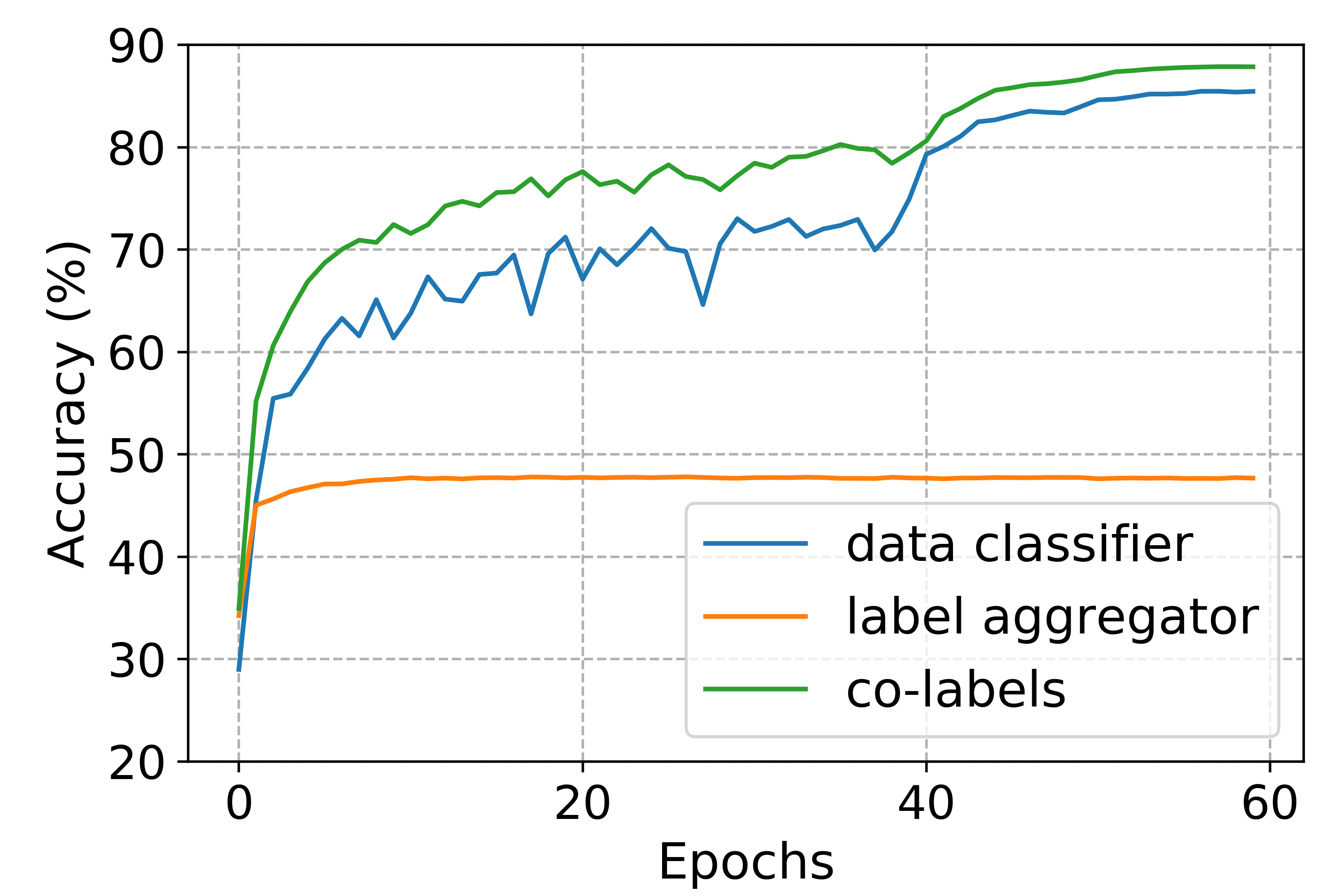}}
	\caption{Accuracy of the data classifier, label aggregator and co-labels on our TCL approach without calibration~(a) and TCL~(b) during iterative training on the IND-3 dataset. }
	\label{ind_3}
	\vspace{-10pt}
\end{figure}

\begin{figure}[h]
	\centering
	\subfloat[]{\includegraphics[width=1.7in]{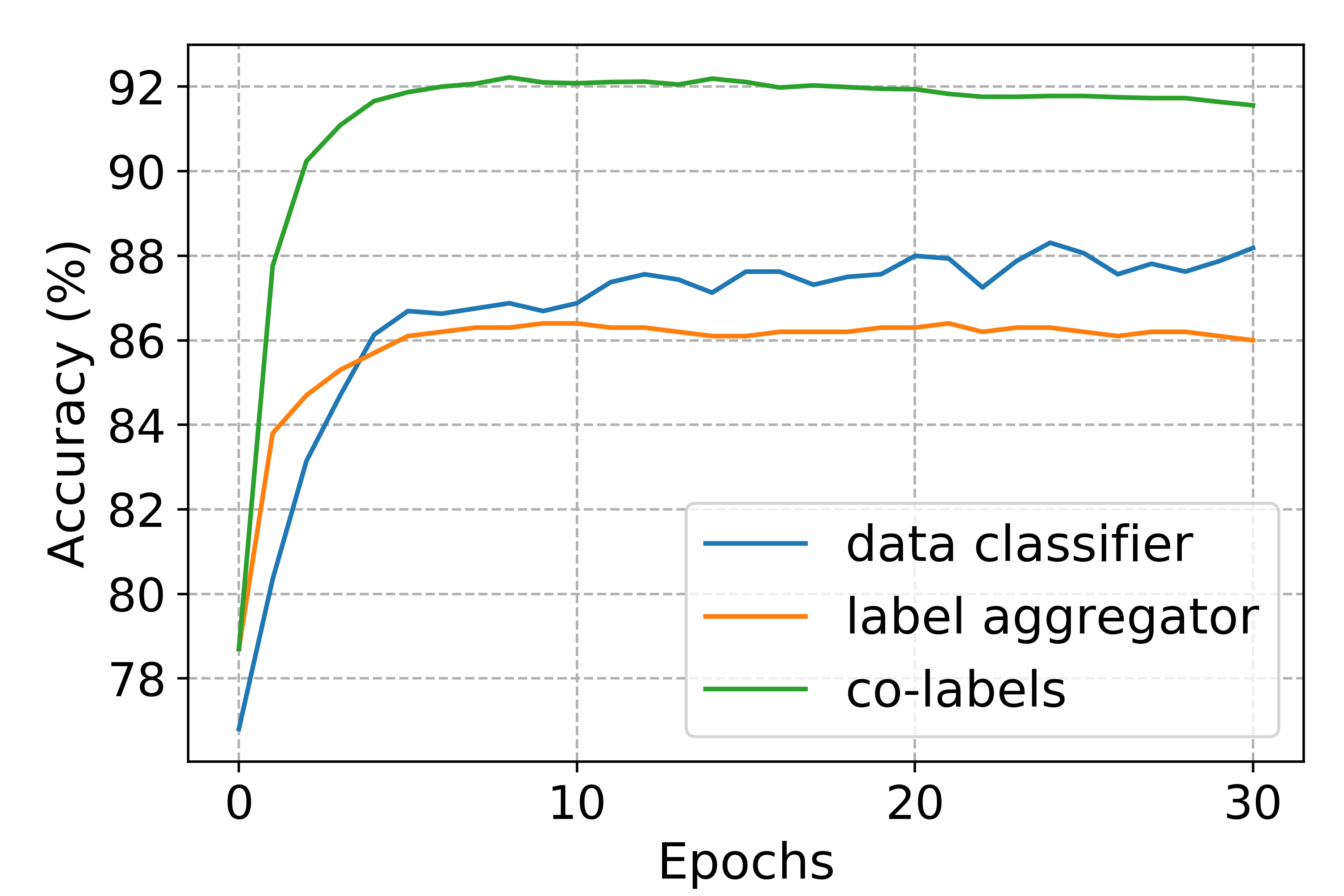}\label{labelme}}
	\hfil
	\subfloat[]{\includegraphics[width=1.7in]{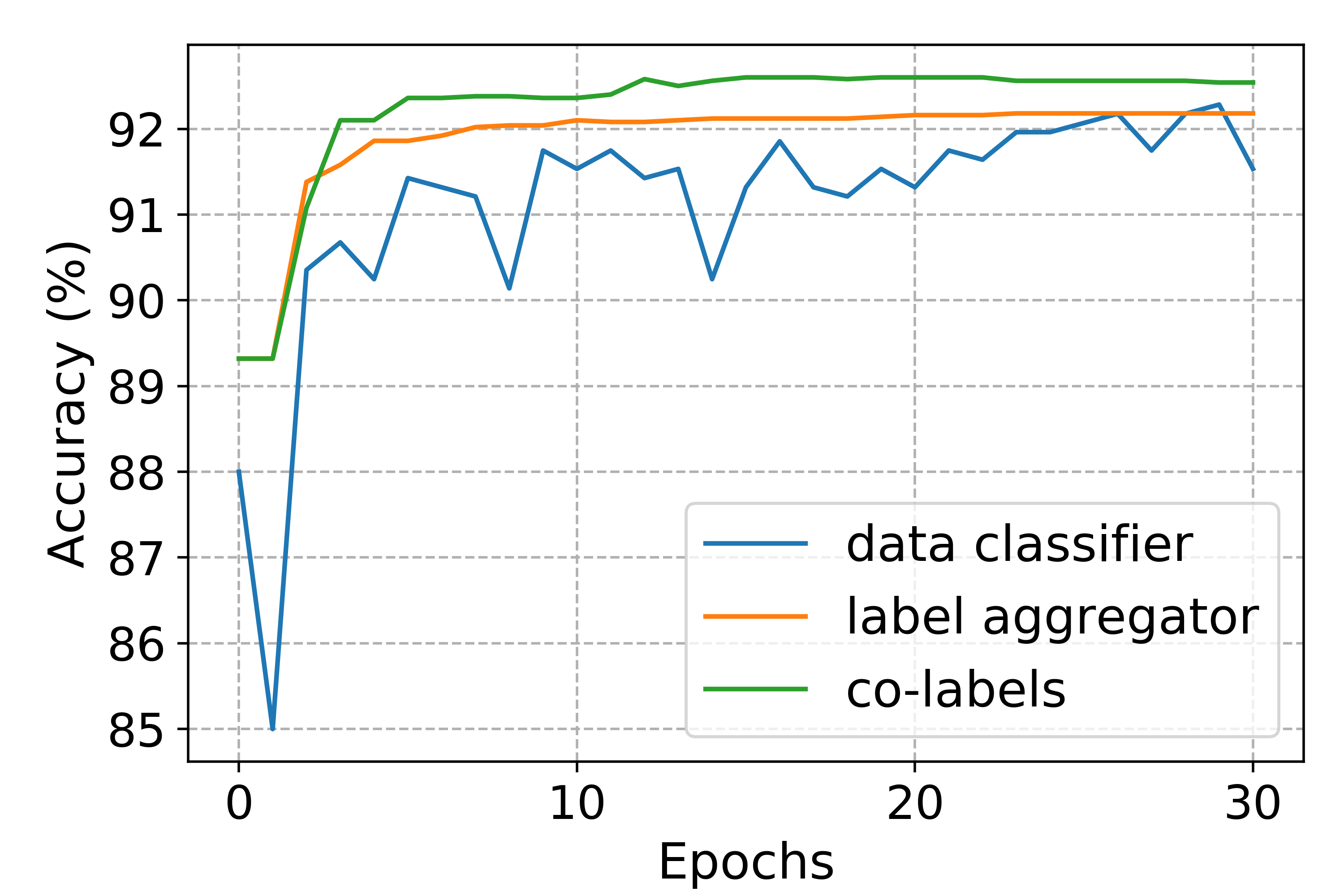}\label{cub}}
	\caption{Accuracy of the data classifier, label aggregator and co-labels on our TCL approach during iterative training on the Labelme-AMT dataset~(a) and CUBShape dataset~(b). }
\end{figure}

\begin{figure}[!h]
	\centering
	\includegraphics[width=1.0\linewidth]{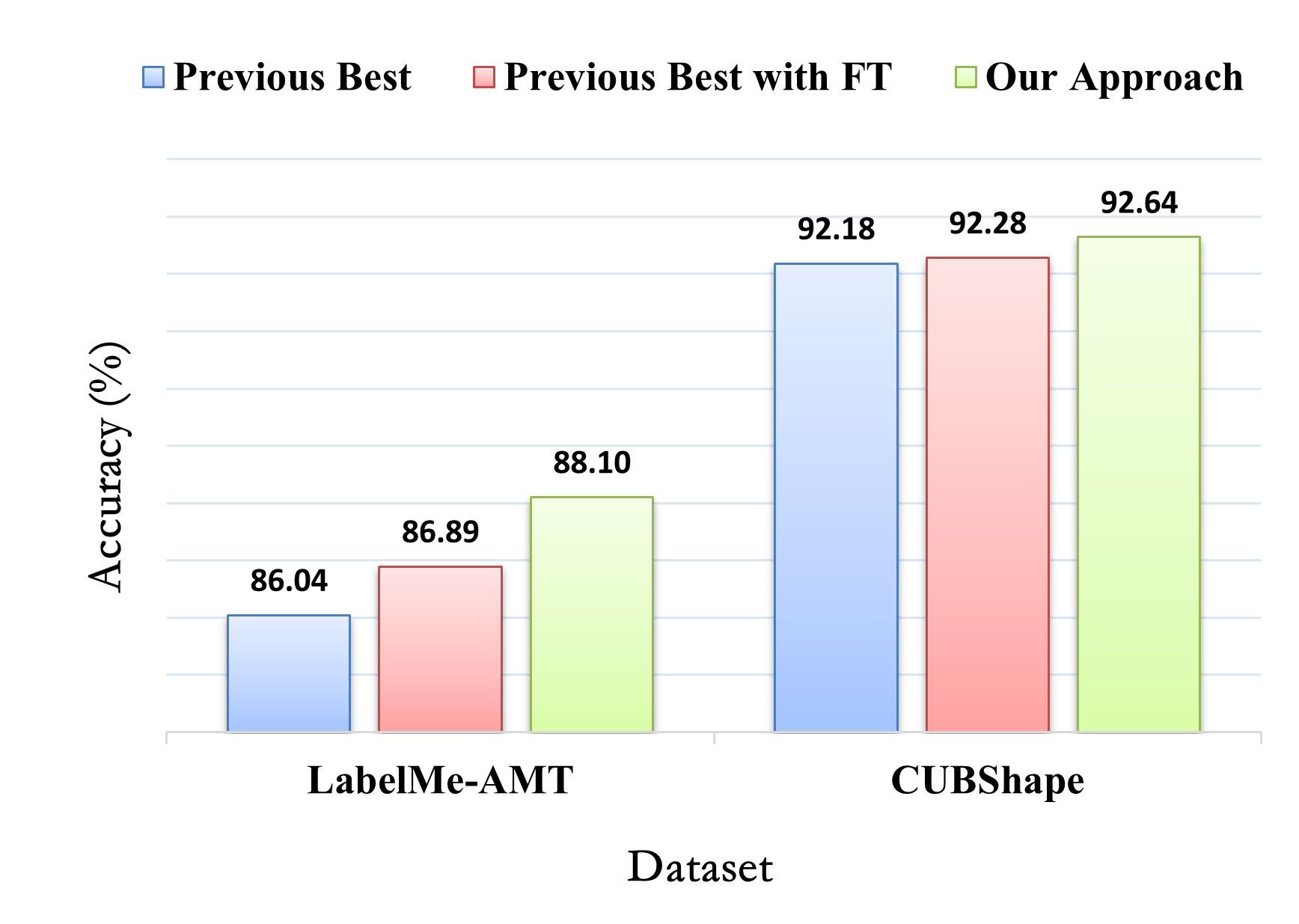}
	\caption{Performance on two real-world crowdsourcing datasets LabelMe-AMT and CUBShape. \textbf{Previous Best} denotes the best results for previous methods without using trusted data. \textbf{Previous Best with FT} denotes the best results for previous methods with the fine-tuning on trusted data. Due to the efficient utilization of the small, trusted dataset, \textbf{Our Approach} substantially outperforms the other state-of-the-art methods.}
	\label{motivation2}
\end{figure}

After the promise is achieved 
on synthetic datasets, we check the performance on the real crowdsourcing dataset LabelMe-AMT. As reported in column 2 of Tab. \ref{labelme_result}, fine-tuning pretrained baselines on trusted data can improve the best accuracy from 86.04\% to 86.89\% (increased by 0.85\%), while our approach achieves a much better accuracy of 88.09\%, with a 2.06\% accuracy improvement compared with the best method without trusted data, demonstrating its superiority in the efficient utilization of the small, trusted dataset. Fig. \ref{labelme} clearly shows the mutual improvement of classifier learning and prediction combination during training on LabelMe-AMT. 
Therefore, not only is deep classifier learned well through such iterative learning, but so is the Naive Bayes classifier, which can estimate the confusion matrices of different annotators more accurately than the other state-of-the-art methods~(see Fig. \ref{labelme_matrices}).

	We further evaluate the performance on another real crowdsourcing dataset, CUBShape (see column 3 of Tab.~\ref{labelme_result}). As reported, since the label noise of this dataset is not severe, the DL-MV method also attains a good result, with a validation accuracy of 91.42\%, and only a 0.10\% improvement (from 92.18\% to 92.28\%) in the best accuracy is obtained by the fine-tuning strategy. However, compared with other methods, our approach still achieves substantial label noise performance gains (0.46\% above 92.18\%), which succeeds in performing progressive improvement via trustable co-label updating~(see Fig.~\ref{cub}). Fig.~\ref{motivation2} shows the performance advantage of our TCL approach on the above two real-world crowdsourcing datasets more clearly.

\begin{table*}[t]

	\small

	\caption{The generated datasets by simulating different correlated annotators on CIFAR10.}

	\centering

	\begin{tabular}{cccccc}

		\hline Datasets & 1st  annotators & 2nd  annotators & 3rd annotators  & 4th annotators & 5th annotators\\

		\hline

		COR-1 & Symmetry, $\varepsilon=0.5$ & Symmetry, $\varepsilon=0.85$ &  Imitative, 1st  & Imitative, 1st & Supportive, 1st \\

		COR-2 & Symmetry, $\varepsilon=0.8$ & Symmetry, $\varepsilon=0.45$ & Imitative, 1st & Opposite, 2nd & Supportive, 2nd\\

		COR-3 & Class-wise, correct class 1 & Symmetry, $\varepsilon=0.55$ & Imitative, 2nd &  Supportive, 2nd &  Opposite, 2nd \\

		COR-4 & Class-wise, correct class 1 & Symmetry, $\varepsilon=0.6$ & Symmetry, $\varepsilon=0.6$ & Supportive, 2nd & Supportive, 3rd \\

		\hline

	\end{tabular}

	\label{group2}

\end{table*}
	
	\subsection{Learning from {Auto-labeling Sources}}
	As mentioned before, the deep learning community has increasing interest in using various automatic labeling sources to collect large-scale labeled datasets, and we can make all training instances labeled by each annotator, which enables powerful modeling methods such as multilayer neural networks. Thus, we improve TCL for the complete data case in this setting. We conduct the experiments on one synthetic dataset (CIFAR10~\cite{Krizhevsky09tr}) and one real dataset (SocialEvent) to evaluate the performance of TCL and TCLS when all training samples are labeled by every auto-labeling source.
	
	\myPara{Datasets.}
	\textbf{CIFAR10} is also used in this case. To verify the effectiveness of label aggregator modeling capacity in our approach, we try to simulate the complex correlation between annotators in real situations and design 4 sets of correlated annotators in this setting (see Tab. \ref{group2}). Each set includes 5 annotators (each annotator labels all training data), and the kinds of correlated label noise include 1) \textbf{imitative labeling}, which labels the instances by the same results as another annotator; 2) \textbf{supportive labeling}, which gives correct labels to the instances that are correctly labeled by another annotator but gives incorrect labels uniformly at random to the remaining instances; and 3) \textbf{opposite labeling}, which gives correct labels to the instances that are incorrectly labeled by another annotator but gives incorrect labels uniformly at random to the remaining instances. We preprocess the images in the same way as the other setting.
	
	\textbf{SocialEvent} is a real large-scale binary classification dataset collected from social media, which is used to predict whether the event is abnormal or not. Each instance includes 134-dimensional preprocessed feature data and 25 noisy labels from automatic labeling sources (e.g., rules, related records, and weak models). We use 470K untrusted data, 300 trusted data and 10K validation data. \textbf{SocialEvent} is a highly class-imbalanced dataset (85$\%$, 15$\%$).

	\myPara{Implementation.}
	For synthetic CIFAR10 datasets, we adopt the VGG-16 networks to model the data classifier. Our TCL approach trains two classifiers for $T=51$ iterations and then retrains the data classifier for 60 epochs.
	In the first iteration round, the data classifier is trained for 10 epochs, and in the other round, it is trained for 1 epoch. During the alternate optimization stage and retraining stage, we use SGD with a batch size of 128, a momentum of 0.9, a weight decay of 0.0005, and an initial learning rate of 0.1. The learning rate of the data classifier network is divided by 10 after 40 epochs and 50 epochs (for a total of 60 epochs). Our TCLS approach uses three-layer fully connected neural networks to model label aggregators, where the first and second hidden layers have 64 and 32 units, respectively (with ReLU activations), and one softmax output layer is on the top. TCLS trains two classifiers for $T=11$ iterations and then retrains the data classifier for 60 epochs. In the first iteration round, the data classifier is trained for 10 epochs, and in the other round, it is trained for 5 epochs. The label aggregator is trained for 3 epochs in each iteration. The optimizer setting for the data classifier is the same as TCL, and for the label aggregator, we use the Adam optimizer with a batch size of 128, a learning rate of 0.001, and betas of (0.9, 0.999).
	
	For the real SocialEvent dataset, for generality, we use three-layer fully connected neural networks to represent the data classifier, where the first and second hidden layers have 128 and 32 units, respectively (with ReLU activations), and one softmax output layer is on the top. Our TCL approach trains two classifiers for $T=10$ iterations and retrains the data classifier for 30 epochs. In each iteration round, the data classifier is trained for 5 epochs. We use the Adam optimizer with a batch size of 128, a learning rate of 0.001, and betas (0.9, 0.999). Our TCLS approach uses the same setting as TCL, except that the label aggregator is modeled as a three-layer fully connected neural network, where the first and second hidden layers have 64 and 32 units, respectively, and we use the Adam optimizer with a batch size of 128, a learning rate of 0.001, and betas (0.9, 0.999) for the aggregator.
	
	\myPara{Results.}~We compare our approaches TCL and TCLS with the same baselines as the other setting. In addition, as the trusted data in this case include noisy labels, we also validate some baselines with initializing parameters or selecting annotators by such trusted data.

	\begin{table}[t]

	\small

	\caption{Validation accuracy ($\%$) on synthetic CIFAR10 datasets with correlated annotators. "F" denotes fine-tuning on trusted data. "I" denotes initializing parameters by trusted data. The minimal improvement is also given.}

	\centering

	\begin{tabular}{p{2.5cm}<{\centering}p{1.1cm}<{\centering}p{1.1cm}<{\centering}p{1.1cm}<{\centering}p{1.1cm}<{\centering}}

		\hline Approach & COR-1  & COR-2 & COR-3 & COR-4 \\

		\hline 	DL-MV &  83.83 & 58.67 & 84.31 & 85.30  \\

		{DL-DS} &  {84.04} & {46.54} & {82.10} & {86.20}  \\

		{DL-IBCC} &  {84.29} & {45.24} & {81.98} & {86.28}  \\

		{DL-CRH} &  {84.25} & {46.99} & {82.62} & {84.99}  \\

		AggNet &  84.80 & 85.81 & 83.54 & 86.87 \\	
		CrowdLayer &  81.41 & 82.39 & 80.43 & 82.70 \\
		MBEM &  84.22 & 85.93 & 83.41 & 85.42   \\

		Max-MIG &  84.27 & 86.40 & 83.53 & 85.58  \\

	    CVL &  84.04 & 85.65 & 84.27 & 86.58  \\

		\hline  DL-MV+F &  83.93 & 77.12 & 84.35 & 85.38  \\

		{DL-DS+F} &  {84.04} & {53.24} & {82.10} & {86.20}   \\

        {DL-IBCC+F} &  {84.29} & {52.54} & {81.98} & {86.28}  \\

        {DL-CRH+F} &  {84.28} & {54.38} & {82.70} & {84.99}  \\		

		AggNet+F &  84.81 & 86.03 & 83.75 & 87.00 \\	 	

		CrowdLayer+F &  81.46 & 82.59 & 80.63 & 82.73 \\
		MBEM+F &  84.25 & 85.99 & 83.43 & 85.56  \\

		Max-MIG+F &  84.27 & 86.50 & 83.65 & 85.64 \\

		CVL+F &  84.33 & 85.92 & 84.49 & 86.63 \\

		\hline

		AggNet+I+F &  84.43 & 86.35 & 84.29 & 86.95 \\

		CrowdLayer+I+F &  80.15 & 82.57 & 69.76 & 82.57 \\

		Max-MIG+I+F &  84.31 & 86.22 & 83.40 & 86.22 \\

		CVL+I+F &  84.32 & 86.40 & \underline{84.79} & 86.51  \\

		\hline 	Our TCL &  \underline{84.89} & 87.39 & 83.68 & 87.45 \\

		Our TCL+I &  84.31 & \underline{90.37} & 83.82 & \underline{88.53}\\

		Our TCLS &  \textbf{90.62} & \textbf{93.09} & \textbf{93.11} & \textbf{91.62} \\

		\hline 	        Min$\uparrow$  &  5.73 & 2.72 & 8.32 & 3.09  \\

		\hline

	\end{tabular}

	\label{cifar10_cor}

    \end{table}

\begin{figure}[h]
	\centering
	\subfloat[]{\includegraphics[width=1.7in]{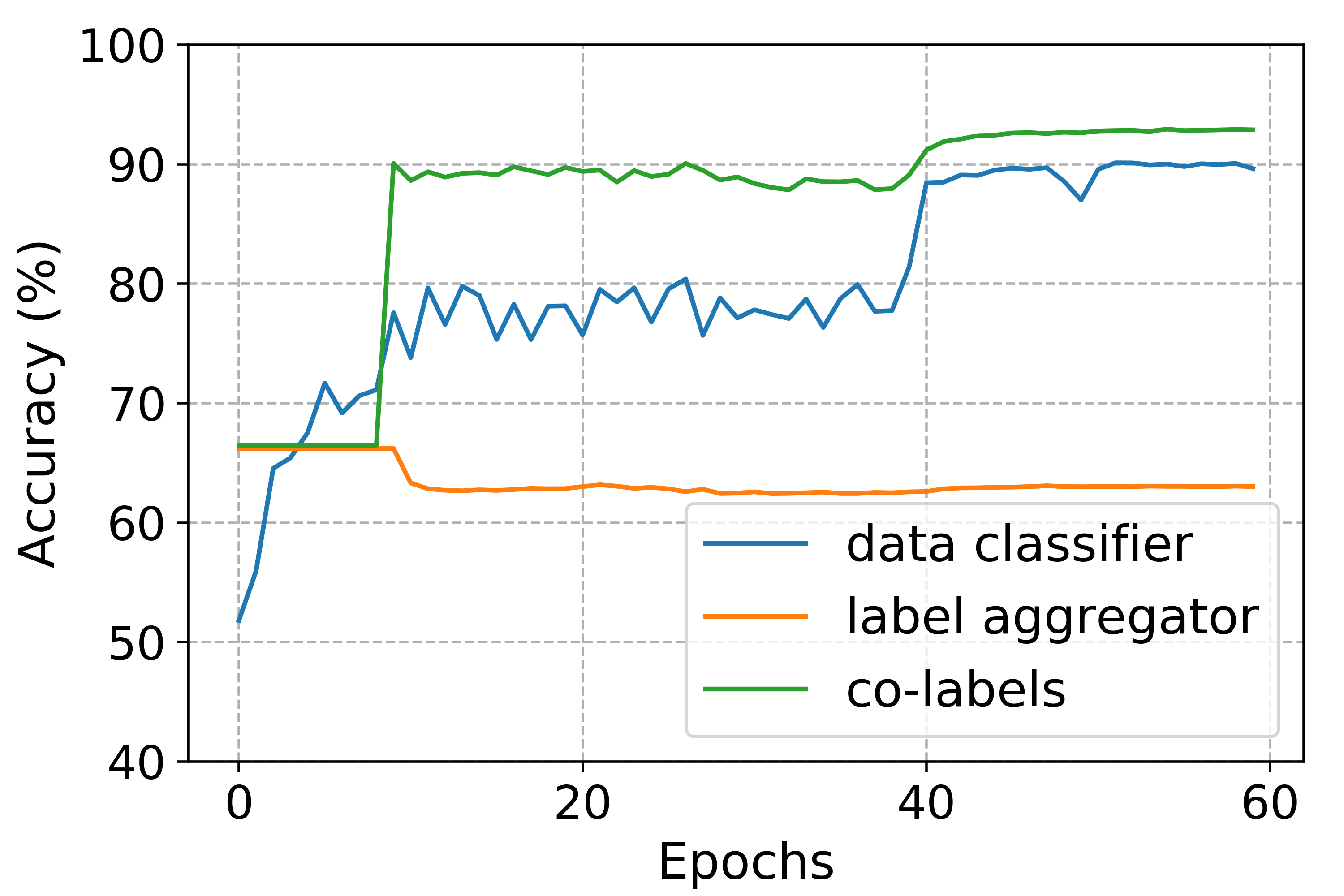}}
	\hfil
	\subfloat[]{\includegraphics[width=1.7in]{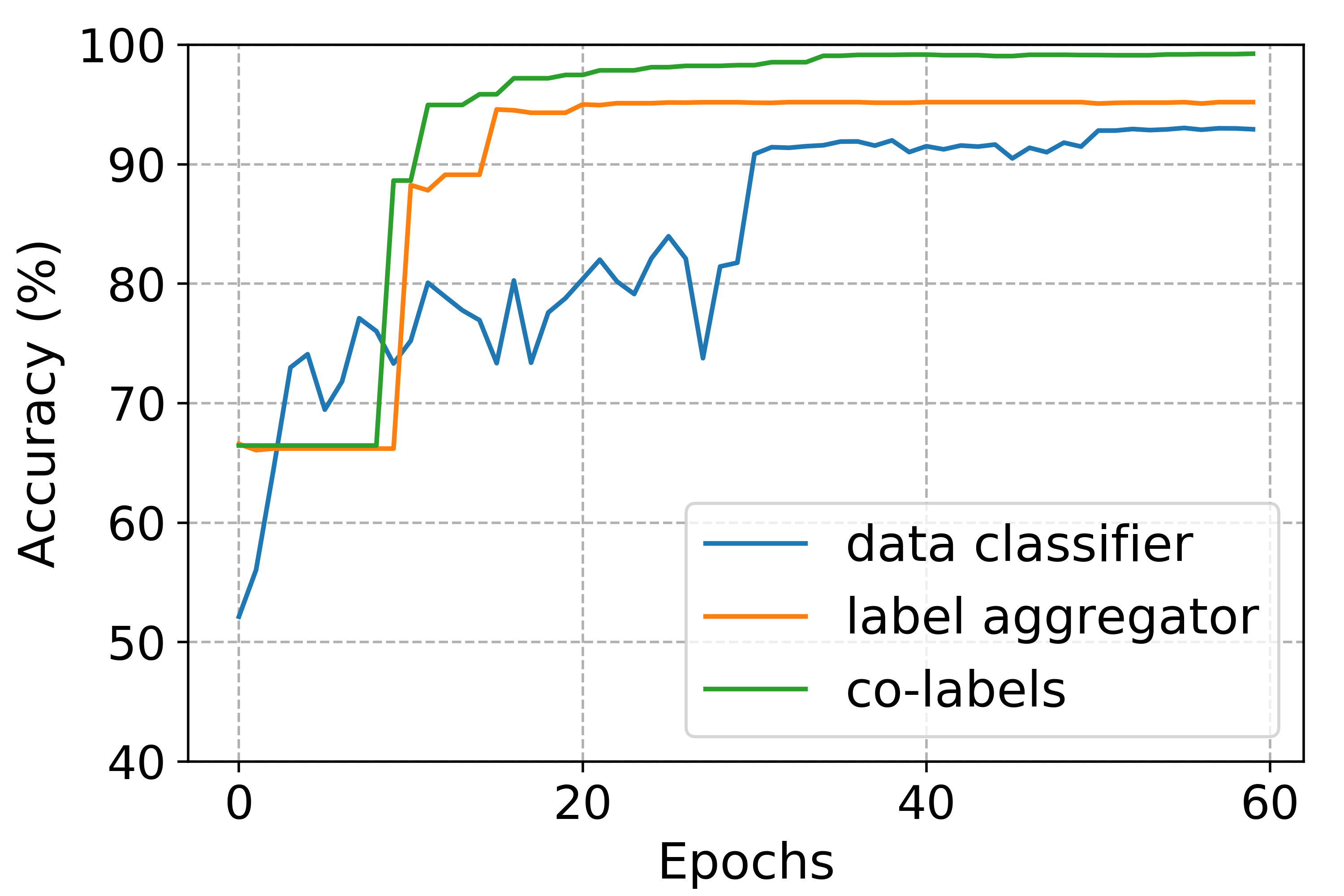}}
	\caption{Accuracy of the data classifier, label aggregator and co-labels on our approaches TCL+I (a) and TCLS (b) during iterative training on the COR-2 dataset.}
	\label{cor-2}
\end{figure}

	\begin{table}[ht]
	\small
	\caption{Validation AUC on SocialEvent dataset. "F" denotes fine-tuning on trusted data. "I" denotes initializing parameters or selecting annotators by trusted data.}

	\centering

	\begin{tabular}{cc}

		\hline Approach  & Validation AUC(\%)  \\

		\hline {DL-MV+I} & {51.25 ($\pm$ 2.64)}\\

		{DL-DS+I} & {51.55 ($\pm$ 0.45)}\\

		{DL-IBCC+I} & {51.09 ($\pm$ 1.35)} \\

		{DL-CRH+I} & {55.07 ($\pm$ 3.05)}\\

		AggNet+I &  62.55 ($\pm$ 0.17)\\

		CrowdLayer+I &  60.44 ($\pm$ 0.25)\\

		MBEM+I &  62.91 ($\pm$ 0.22)\\

		MAX-MIG+I &  56.22 ($\pm$ 1.45)\\

		CVL+I &  63.88 ($\pm$ 0.07)\\

		\hline {DL-MV+I+F} & {63.20 ($\pm$ 1.05)}\\

		{DL-DS+I+F} & {70.73 ($\pm$ 1.73)}\\

		{DL-IBCC+I+F} & {73.79 ($\pm$ 1.07)}\\

		{DL-CRH+I+F} & {72.57 ($\pm$ 1.79)}\\		

		AggNet+I+F & 66.70 ($\pm$ 3.75)\\

		CrowdLayer+I+F & 72.81 ($\pm$ 1.58)\\

		MBEM+I+F &  73.18 ($\pm$ 0.78)\\

		MAX-MIG+I+F &  61.73 ($\pm$ 0.88)\\

		CVL+I+F &  73.00 ($\pm$ 0.97)\\

		\hline Our TCL &  69.14 ($\pm$ 6.38)\\

		Our TCL+I &  74.05 ($\pm$ 0.69)\\

		Our TCLS &  \textbf{75.21 ($\pm$ 0.24)}\\

		\hline

	\end{tabular}

	\label{credit_result}

\end{table}

	The results on four synthetic CIFAR10 datasets under this setting are shown in Tab. \ref{cifar10_cor}. Obviously, we can find that our TCLS delivers much better accuracy than other methods on all four noisy datasets. The average minimal improved accuracy reaches 4.97\% on four datasets, which empirically demonstrates the advantage of modeling correlated annotators by neural networks. Fig. \ref{cor-2} shows the progressive learning process of TCL+I and TCLS during training on the COR-2 dataset, which clearly shows the huge performance gap between the label aggregator modeled by neural networks and by the Naive Bayes classifier, and this also further leads to the much superior learning effectiveness of TCLS.
	
	In addition
, we can obtain some other observations from these results. First, the correlated annotators make the labeling assumption of other state-of-the-art methods~(including TCL) not hold, and therefore, their performance is unstable on such four datasets, which means those methods perform well on one dataset, but poorly on another dataset; for example, on COR-3 dataset, most approaches are inferior to the simple DL-MV approach with the fine-tuning, which achieves a validation accuracy of 84.35\%. Second, similar to the performance on generated crowdsourcing datasets, a simple fine-tuning strategy has a limited effect on improving the good pretrained baselines.
	Third, initializing confusion matrices or pseudo labels via trusted data for baselines does not always work well on these four datasets, which may be because all annotators are positive and initialization by majority voting is enough at most times.

	\begin{figure*}[!ht]
	\centering
	\subfloat[]{\includegraphics[width=1.75in]{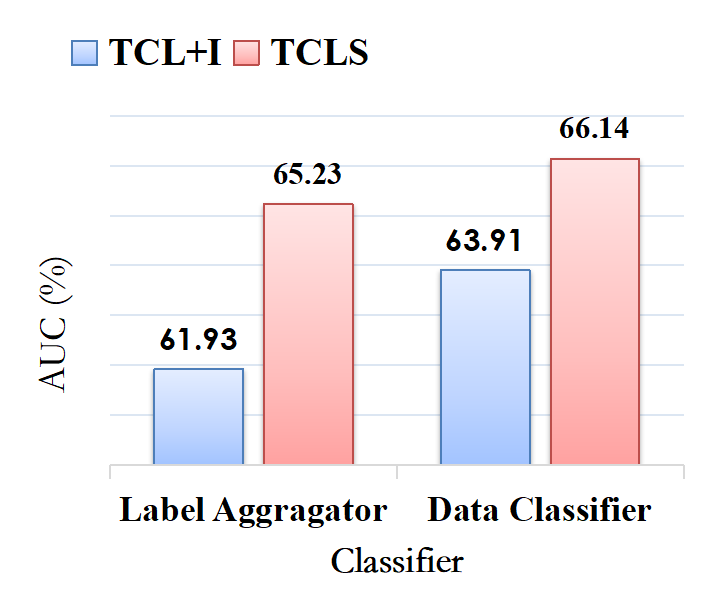}\label{SocialEvent}}
	\hfil
	\subfloat[]{\includegraphics[width=1.75in]{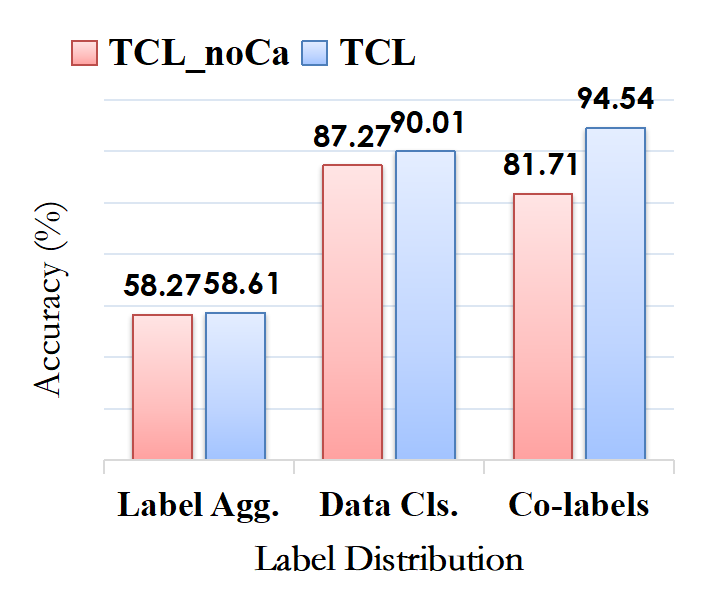}\label{calibration}}
	\hfil
	\subfloat[]{\includegraphics[width=1.75in]{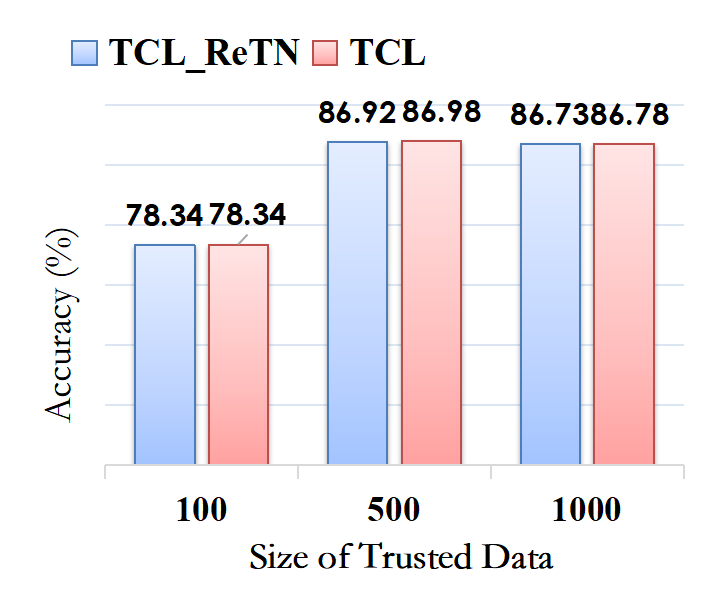}\label{size_of_trusted_data}}
	\hfil
	\subfloat[]{\includegraphics[width=1.75in]{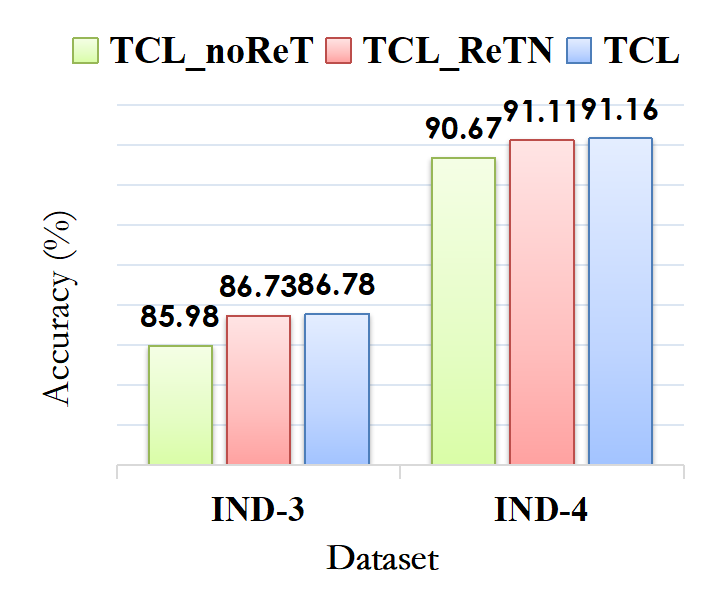}\label{re-training}}
	\caption{(a) AUC of classifiers on TCL+I and TCLS after alternate optimization on the SocialEvent dataset; (b) Impact of calibration on TCL. (c) Impact of size of trusted data on TCL. (d) Impact of retraining strategy on TCL.}
\end{figure*}	
	
	
	We also evaluate the performance on the real dataset SocialEvent with multiple auto-labeling sources. The results are shown in Tab. \ref{credit_result}. {The SocialEvent dataset has many adversaries who deliberately give many incorrect labels, and their labeling accuracies are even lower than the prior probability of the negative class (85\%).} {Therefore, those methods without any trusted data all fail, and we only present the results with initializing parameters or selecting annotators by trusted data.} Faced with such heavy label noise, a simple fine-tuning technique also achieves great performance improvement. In addition, between those methods using trusted data, TCLS achieves better performance, with an AUC of 75.21\%, again verifying its advantages compared with the other state-of-the-art methods. Fig. \ref{SocialEvent} shows the validation AUC of two classifiers in TCL+I and TCLS after alternate optimization, from which we can verify that the performance of the label aggregator benefits from the high modeling capacity of deep networks during iteration, also making the data classifier learning more effective.

	\subsection{Ablation Study}
	\label{ablation}
	We mainly study the impact of different components on TCL, and the results on TCLS are similar.

	\myPara{Impact of calibration.}
	To study the impact of calibration, we use the four generated CIFAR10 datasets with independent annotators to perform the evaluation. Fig. \ref{calibration} shows the average accuracy of our TCL approach with and without calibration~($\rm{TCL_{noCa}}$) after the alternate optimization stage on four datasets, { which verifies the claim that calibration via trusted data plays a critical role in guiding the trustable prediction combination and that our approach utilizes trusted data effectively.}
	
	\myPara{Size of trusted data.}
	Trusted data provide important extra information for TCL to guide the learning process. To study the impact of the size of trusted data, we validate the performance of TCL under different sizes of trusted data on the IND-3 dataset. To decouple the effect of calibration and clean sample training, we also validate the performance of TCL with retraining only on noisy data~($\rm{TCL_{ReTN}}$).
	As shown in Fig. \ref{size_of_trusted_data}, although our approach is more data-efficient than the fine-tuning technique, the performance of our approach will decline if the size of trusted data is too small.

\myPara{Impact of retraining strategy.}
	To study the impact of the retraining strategy on generated crowdsourcing datasets IND-3 and IND-4, we compare the performance of TCL, TCL with retraining on noisy data with co-labels~($\rm{TCL_{ReTN}}$), and TCL without retraining~($\rm{TCL_{noReT}}$).
	The results in Fig. \ref{re-training} show that the retraining strategy further effectively improves the final performance of the data classifier, and both untrusted data and trusted data make positive contributions to the retraining stage.

\myPara{Impact of co-label initialization on TCLS.}
	Our TCLS approach uses trusted data to train a Naive Bayes classifier to initialize co-labels. To study this effect, we conduct majority voting to initialize co-labels and perform experiments on COR-2 and SocialEvent datasets. After alternate optimization, TCLS achieves an accuracy of 93.04\% on the COR-2 dataset, and TCLS with co-label initialization by majority voting achieves an accuracy of 92.87\%. TCLS attains an AUC of 66.13\% on the SocialEvent dataset, while TCLS with co-label initialization by majority voting attains an AUC of 63.41\%. These results are consistent with other baselines, which shows that having good initialization is important when there are many adversaries in annotators, and initialization by majority voting is not undesirable in the usual case.

	\section{Conclusion}
{In this paper, we propose a \emph{Trustable Co-label Learning} approach to make full use of a small amount of trusted data to facilitate robust classifier learning from multiple annotators.} It alternately refines classifiers and relabels pseudo labels with the guidance of a small trusted dataset based on the conditionally independent assumption. In addition, {we further improve this approach by modeling the label aggregator as a multilayer neural network to enhance modeling ability when all instances are labeled by every annotator.} The proposed approach achieves superior performance in terms of effectiveness and robustness on both synthetic and real datasets.
	In the future, we will explore a more complicated but still trainable probabilistic aggregator modeling way to represent the dependence among different annotators when there are missing labels.

\myPara{Acknowledgements}. This work was partially supported by grants from the National Key Research and Development Plan (2020AAA0140001), National Natural Science Foundation of China (61772513), and Beijing Natural Science Foundation (19L2040). Shiming Ge is also supported by the Youth Innovation Promotion Association, Chinese Academy of Sciences.

\bibliographystyle{IEEEtran}
\bibliography{bibTCL}

\begin{thebibliography}{10}
\providecommand{\url}[1]{#1}
\csname url@samestyle\endcsname
\providecommand{\newblock}{\relax}
\providecommand{\bibinfo}[2]{#2}
\providecommand{\BIBentrySTDinterwordspacing}{\spaceskip=0pt\relax}
\providecommand{\BIBentryALTinterwordstretchfactor}{4}
\providecommand{\BIBentryALTinterwordspacing}{\spaceskip=\fontdimen2\font plus
\BIBentryALTinterwordstretchfactor\fontdimen3\font minus
  \fontdimen4\font\relax}
\providecommand{\BIBforeignlanguage}[2]{{%
\expandafter\ifx\csname l@#1\endcsname\relax
\typeout{** WARNING: IEEEtran.bst: No hyphenation pattern has been}%
\typeout{** loaded for the language `#1'. Using the pattern for}%
\typeout{** the default language instead.}%
\else
\language=\csname l@#1\endcsname
\fi
#2}}
\providecommand{\BIBdecl}{\relax}
\BIBdecl

\bibitem{he2016deep}
K.~He, X.~Zhang, S.~Ren, and J.~Sun, ``Deep residual learning for image
  recognition,'' in \emph{IEEE Conference on Computer Vision and Pattern
  Recognition (CVPR)}, 2016, pp. 770--778.

\bibitem{ren2015faster}
S.~Ren, K.~He, R.~Girshick, and J.~Sun, ``Faster r-cnn: Towards real-time
  object detection with region proposal networks,'' in \emph{Neural Information
  Processing Systems (NeurIPS)}, 2015, pp. 91--99.

\bibitem{ChenHNLLWX18}
H.~Chen, F.~X. Han, D.~Niu, D.~Liu, K.~Lai, C.~Wu, and Y.~Xu, ``{MIX:}
  multi-channel information crossing for text matching,'' in \emph{ACM SIGKDD
  Conference on Knowledge Discovery and Data Mining (KDD)}, 2018, pp. 110--119.

\bibitem{Sun2018DPPDL}
Z.~Sun, Q.~Zhang, Y.~Li, and Y.~Tan, ``Dppdl: A dynamic partial-parallel data
  layout for green video surveillance storage,'' \emph{IEEE Transactions on
  Circuits and Systems for Video Technology}, vol.~28, no.~1, pp. 193--205,
  2018.

\bibitem{WangLLXRZMLHZ20}
G.~Wang, X.~Liu, C.~Li, Z.~Xu, J.~Ruan, H.~Zhu, T.~Meng, K.~Li, N.~Huang, and
  S.~Zhang, ``A noise-robust framework for automatic segmentation of {COVID-19}
  pneumonia lesions from {CT} images,'' \emph{{IEEE} Transactions on Medical
  Imaging}, vol.~39, no.~8, pp. 2653--2663, 2020.

\bibitem{YaoZSHXT17}
Y.~Yao, J.~Zhang, F.~Shen, X.~Hua, J.~Xu, and Z.~Tang, ``Exploiting web images
  for dataset construction: {A} domain robust approach,'' \emph{IEEE
  Transactions on Multimedia}, vol.~19, no.~8, pp. 1771--1784, 2017.

\bibitem{WangWJ14}
H.~Wang, X.~Wu, and Y.~Jia, ``Video annotation via image groups from the web,''
  \emph{IEEE Transactions on Multimedia}, vol.~16, no.~5, pp. 1282--1291, 2014.

\bibitem{ChaudharyGPC20}
C.~Chaudhary, P.~Goyal, D.~N. Prasad, and Y.~P. Chen, ``Enhancing the quality
  of image tagging using a visio-textual knowledge base,'' \emph{IEEE
  Transactions on Multimedia}, vol.~22, no.~4, pp. 897--911, 2020.

\bibitem{9113752}
Y.~{Lin} and H.~{Chen}, ``Tag propagation and cost-sensitive learning for music
  auto-tagging,'' \emph{IEEE Transactions on Multimedia}, vol.~23, pp.
  1605--1616, 2021.

\bibitem{abs-1708-02862}
W.~Li, L.~Wang, W.~Li, E.~Agustsson, and L.~V. Gool, ``Webvision database:
  Visual learning and understanding from web data,'' \emph{CoRR
  abs/1708.02862}, 2017.

\bibitem{Zhang2017iclr}
C.~Zhang, S.~Bengio, M.~Hardt, B.~Recht, and O.~Vinyals, ``Understanding deep
  learning requires rethinking generalization,'' in \emph{International
  Conference on Learning Representations (ICLR)}, 2017.

\bibitem{Liu2016TPAMI}
T.~Liu and D.~Tao, ``Classification with noisy labels by importance
  reweighting,'' \emph{IEEE Transactions on Pattern Analysis and Machine
  Intelligence}, vol.~38, no.~3, pp. 447--461, 3 2016.

\bibitem{cheng2019learning}
J.~Cheng, T.~Liu, K.~Ramamohanarao, and D.~Tao, ``Learning with bounded
  instance-and label-dependent label noise,'' in \emph{International Conference
  on Machine Learning (ICML)}, 2020, pp. 1789--1799.

\bibitem{XiaLW00NS19}
X.~Xia, T.~Liu, N.~Wang, B.~Han, C.~Gong, G.~Niu, and M.~Sugiyama, ``Are anchor
  points really indispensable in label-noise learning?'' in \emph{Advances in
  Neural Information Processing Systems (NeurIPS)}, 2019, pp. 6835--6846.

\bibitem{DengDSLL009}
J.~Deng, W.~Dong, R.~Socher, L.~Li, K.~Li, and F.~Li, ``Imagenet: {A}
  large-scale hierarchical image database,'' in \emph{{IEEE} Computer Society
  Conference on Computer Vision and Pattern Recognition (CVPR)}, 2009, pp.
  248--255.

\bibitem{BielG13}
J.~Biel and D.~Gatica{-}Perez, ``The youtube lens: Crowdsourced personality
  impressions and audiovisual analysis of vlogs,'' \emph{{IEEE} Transactions on
  Multimedia}, vol.~15, no.~1, pp. 41--55, 2013.

\bibitem{ServajeanJSCP17}
M.~Servajean, A.~Joly, D.~E. Shasha, J.~Champ, and E.~Pacitti, ``Crowdsourcing
  thousands of specialized labels: {A} bayesian active training approach,''
  \emph{{IEEE} Transactions on Multimedia}, vol.~19, no.~6, pp. 1376--1391,
  2017.

\bibitem{WhitehillRWBM09}
J.~Whitehill, P.~Ruvolo, T.~Wu, J.~Bergsma, and J.~R. Movellan, ``Whose vote
  should count more: Optimal integration of labels from labelers of unknown
  expertise,'' in \emph{Neural Information Processing Systems (NeurIPS)}, 2009,
  pp. 2035--2043.

\bibitem{LiLGZFH14}
Q.~Li, Y.~Li, J.~Gao, B.~Zhao, W.~Fan, and J.~Han, ``Resolving conflicts in
  heterogeneous data by truth discovery and source reliability estimation,'' in
  \emph{International Conference on Management of Data (SIGMOD)}, 2014, pp.
  1187--1198.

\bibitem{Dawid1979Maximum}
A.~P. Dawid and A.~M. Skene, ``Maximum likelihood estimation of observer
  error-rates using the em algorithm,'' \emph{Journal of the Royal Statistical
  Society}, vol.~28, no.~1, pp. 20--28, 1979.

\bibitem{KimG12}
H.~Kim and Z.~Ghahramani, ``Bayesian classifier combination,'' in
  \emph{Proceedings of the Fifteenth International Conference on Artificial
  Intelligence and Statistics (AISTATS)}, vol.~22, 2012, pp. 619--627.

\bibitem{Raykar2010JMLR}
V.~C. Raykar, S.~Yu, L.~H. Zhao, G.~H. Valadez, C.~Florin, L.~Bogoni, and
  L.~Moy, ``Learning from crowds,'' \emph{Journal of Machine Learning
  Research}, vol.~11, pp. 1297--1322, 2010.

\bibitem{ZhangSLW18}
J.~Zhang, V.~S. Sheng, T.~Li, and X.~Wu, ``Improving crowdsourced label quality
  using noise correction,'' \emph{IEEE Transactions on Neural Networks and
  Learning Systems}, vol.~29, no.~5, pp. 1675--1688, 2018.

\bibitem{Albarqouni2016TMI}
S.~Albarqouni, C.~Baur, F.~Achilles, and \etal, ``Aggnet: Deep learning from
  crowds for mitosis detection in breast cancer histology images,'' \emph{IEEE
  Transactions on Medical Imaging}, vol.~35, no.~5, pp. 1313--1321, 2016.

\bibitem{Rodrigues2017TPAMI}
F.~Rodrigues, M.~Lourenco, B.~Ribeiro, and F.~Pereira, ``Learning supervised
  topic models for classification and regression from crowds,'' \emph{IEEE
  Transactions on Pattern Analysis and Machine Intelligence}, vol.~39, no.~12,
  pp. 2409--2422, 2017.

\bibitem{YinL0020}
L.~Yin, Y.~Liu, W.~Zhang, and Y.~Yu, ``Aggregating crowd wisdom with side
  information via a clustering-based label-aware autoencoder,'' in
  \emph{International Joint Conference on Artificial Intelligence (IJCAI)},
  2020, pp. 1542--1548.

\bibitem{CaoXKW19}
P.~Cao, Y.~Xu, Y.~Kong, and Y.~Wang, ``Max-mig: an information theoretic
  approach for joint learning from crowds,'' in \emph{International Conference
  on Learning Representations (ICLR)}, 2019.

\bibitem{Khetan2018iclr}
A.~Khetan, A.~Anandkumar, and Z.~C. Lipton, ``Learning from noisy singly
  labeled data,'' in \emph{International Conference on Learning Representations
  (ICLR)}, 2018.

\bibitem{Li2020aaai}
S.~Li, S.~Ge, Y.~Hua, C.~Zhang, H.~Wen, T.~Liu, and W.~Wang, ``Coupled-view
  deep classifier learning from multiple noisy annotators,'' in \emph{AAAI
  Conference on Artificial Intelligence (AAAI)}, 2020, pp. 4667--4674.

\bibitem{yu2010roles}
D.~Yu, L.~Deng, and G.~Dahl, ``Roles of pre-training and fine-tuning in
  context-dependent dbn-hmms for real-world speech recognition,'' in
  \emph{NeurIPS Workshop on Deep Learning and Unsupervised Feature Learning},
  2010.

\bibitem{GuoPSW17}
C.~Guo, G.~Pleiss, Y.~Sun, and K.~Q. Weinberger, ``On calibration of modern
  neural networks,'' in \emph{International Conference on Machine Learning
  (ICML)}, 2017, pp. 1321--1330.

\bibitem{Li2017iccv}
Y.~Li, J.~Yang, and Y.~Song, ``Learning from noisy labels with distillation,''
  in \emph{IEEE International Conference on Computer Vision (ICCV)}, 2017, pp.
  1928--1936.

\bibitem{9337209}
J.~{Chen}, P.~{Ying}, X.~{Fu}, X.~{Luo}, H.~{Guan}, and K.~{Wei}, ``Automatic
  tagging by leveraging visual and annotated features in social media,''
  \emph{IEEE Transactions on Multimedia}, 2021.

\bibitem{DengJTGL14}
C.~Deng, R.~Ji, D.~Tao, X.~Gao, and X.~Li, ``Weakly supervised multi-graph
  learning for robust image reranking,'' \emph{IEEE Transactions on
  Multimedia}, vol.~16, no.~3, pp. 785--795, 2014.

\bibitem{ZhouPBM12}
D.~Zhou, J.~C. Platt, S.~Basu, and Y.~Mao, ``Learning from the wisdom of crowds
  by minimax entropy,'' in \emph{Neural Information Processing Systems
  (NeurIPS)}, 2012, pp. 2204--2212.

\bibitem{BachHRR17}
S.~H. Bach, B.~D. He, A.~Ratner, and C.~R{\'{e}}, ``Learning the structure of
  generative models without labeled data,'' in \emph{International Conference
  on Machine Learning (ICML)}, 2017, pp. 273--282.

\bibitem{ZhouLPM14}
D.~Zhou, Q.~Liu, J.~C. Platt, and C.~Meek, ``Aggregating ordinal labels from
  crowds by minimax conditional entropy,'' in \emph{International Conference on
  Machine Learning (ICML)}, 2014, pp. 262--270.

\bibitem{SimpsonRPS13}
E.~Simpson, S.~J. Roberts, I.~Psorakis, and A.~M. Smith, ``Dynamic bayesian
  combination of multiple imperfect classifiers,'' in \emph{Decision Making and
  Imperfection}, 2013, vol. 474, pp. 1--35.

\bibitem{BiWKT14}
W.~Bi, L.~Wang, J.~T. Kwok, and Z.~Tu, ``Learning to predict from crowdsourced
  data,'' in \emph{Conference on Uncertainty in Artificial Intelligence (UAI)},
  2014, pp. 82--91.

\bibitem{Kurve0K15}
A.~Kurve, D.~J. Miller, and G.~Kesidis, ``Multicategory crowdsourcing
  accounting for variable task difficulty, worker skill, and worker
  intention,'' \emph{IEEE Transactions on Knowledge and Data Engineering},
  vol.~27, no.~3, pp. 794--809, 2015.

\bibitem{VarmaSHRR19}
P.~Varma, F.~Sala, A.~He, A.~Ratner, and C.~R{\'{e}}, ``Learning dependency
  structures for weak supervision models,'' in \emph{International Conference
  on Machine Learning (ICML)}, 2019, pp. 6418--6427.

\bibitem{IpeirotisPSW14}
P.~G. Ipeirotis, F.~J. Provost, V.~S. Sheng, and J.~Wang, ``Repeated labeling
  using multiple noisy labelers,'' \emph{Data Mining and Knowledge Discovery},
  vol.~28, no.~2, pp. 402--441, 2014.

\bibitem{Li014}
H.~Li and B.~Yu, ``Error rate bounds and iterative weighted majority voting for
  crowdsourcing,'' \emph{CoRR abs/1411.4086}, 2014.

\bibitem{ParameswaranSGPW11}
A.~G. Parameswaran, A.~D. Sarma, H.~Garcia{-}Molina, N.~Polyzotis, and
  J.~Widom, ``Human-assisted graph search: it's okay to ask questions,''
  \emph{Proceedings of the VLDB Endowment}, vol.~4, no.~5, pp. 267--278, 2011.

\bibitem{TianZ15}
T.~Tian and J.~Zhu, ``Max-margin majority voting for learning from crowds,'' in
  \emph{Neural Information Processing Systems (NeurIPS)}, 2015, pp. 1621--1629.

\bibitem{KargerOS11}
D.~R. Karger, S.~Oh, and D.~Shah, ``Budget-optimal crowdsourcing using low-rank
  matrix approximations,'' in \emph{Allerton Conference on Communication,
  Control, and Computing (Allerton)}, 2011, pp. 284--291.

\bibitem{DalviDKR13}
N.~N. Dalvi, A.~Dasgupta, R.~Kumar, and V.~Rastogi, ``Aggregating crowdsourced
  binary ratings,'' in \emph{The World Wide Web Conference (WWW)}, 2013, pp.
  285--294.

\bibitem{ZhouH16}
Y.~Zhou and J.~He, ``Crowdsourcing via tensor augmentation and completion,'' in
  \emph{International Joint Conference on Artificial Intelligence (IJCAI)},
  2016, pp. 2435--2441.

\bibitem{Rodrigues2018aaai}
F.~Rodrigues and F.~C. Pereira, ``Deep learning from crowds,'' in \emph{AAAI
  Conference on Artificial Intelligence (AAAI)}, 2018, pp. 1611--1618.

\bibitem{Guan2018aaai}
M.~Y. Guan, V.~Gulshan, A.~M. Dai, and G.~E. Hinton, ``Who said what: Modeling
  individual labelers improves classification,'' in \emph{AAAI Conference on
  Artificial Intelligence (AAAI)}, 2018, pp. 3109--3118.

\bibitem{Chu0W21}
Z.~Chu, J.~Ma, and H.~Wang, ``Learning from crowds by modeling common
  confusions,'' in \emph{{AAAI} Conference on Artificial Intelligence (AAAI)},
  2021, pp. 5832--5840.

\bibitem{Blum1988COLT}
T.~Mitchell and A.~Blum, ``Combining labeled and unlabeled data with
  co-training,'' in \emph{Annual Conference on Learning Theory (COLT)}, 1998,
  pp. 92--100.

\bibitem{Goldman2000ICML}
S.~Goldman and Y.~Zhou, ``Enhancing supervised learning with unlabeled data,''
  in \emph{International Conference on Machine Learning (ICML)}, 2000, pp.
  327--334.

\bibitem{Ma2017icml}
F.~Ma, D.~Meng, Q.~Xie, and \etal, ``Self-paced co-training,'' in
  \emph{International Conference on Machine Learning (ICML)}, 2017, pp.
  2275--2284.

\bibitem{Sindhwani2008icml}
Sindhwani, Vikas, Rosenberg, and S.~David, ``An rkhs for multi-view learning
  and manifold co-regularization,'' in \emph{International Conference on
  Machine Learning (ICML)}, 2008, pp. 976--983.

\bibitem{Ye2015cikm}
H.-J. Ye, D.-C. Zhan, Y.~Miao, and \etal, ``Rank consistency based multi-view
  learning: A privacy-preserving approach,'' in \emph{ACM International
  Conference on Information and Knowledge Management (CIKM)}, 2015, pp.
  991--1000.

\bibitem{lanckriet2004learning}
G.~R. Lanckriet, N.~Cristianini, P.~Bartlett, L.~E. Ghaoui, and M.~I. Jordan,
  ``Learning the kernel matrix with semidefinite programming,'' \emph{Journal
  of Machine learning research}, vol.~5, no. Jan, pp. 27--72, 2004.

\bibitem{cortes2010two}
C.~Cortes, M.~Mohri, and A.~Rostamizadeh, ``Two-stage learning kernel
  algorithms,'' in \emph{International Conference on Machine Learning (ICML)},
  2010, pp. 239--246.

\bibitem{kumar2011co}
A.~Kumar, P.~Rai, and H.~Daume, ``Co-regularized multi-view spectral
  clustering,'' in \emph{Neural Information Processing Systems (NeurIPS)},
  2011, pp. 1413--1421.

\bibitem{domeniconi2007locally}
C.~Domeniconi, D.~Gunopulos, S.~Ma, B.~Yan, M.~Al-Razgan, and D.~Papadopoulos,
  ``Locally adaptive metrics for clustering high dimensional data,'' \emph{Data
  Mining and Knowledge Discovery}, vol.~14, no.~1, pp. 63--97, 2007.

\bibitem{zhou2012ensemble}
Z.-H. Zhou, \emph{Ensemble methods: foundations and algorithms}.\hskip 1em plus
  0.5em minus 0.4em\relax Chapman and Hall/CRC, 2012.

\bibitem{kim2011weight}
H.~Kim, H.~Kim, H.~Moon, and H.~Ahn, ``A weight-adjusted voting algorithm for
  ensembles of classifiers,'' \emph{Journal of the Korean Statistical Society},
  vol.~40, pp. 437--449, 2011.

\bibitem{titterington1981comparison}
D.~Titterington, G.~Murray, L.~Murray, D.~Spiegelhalter, A.~Skene, J.~Habbema,
  and G.~Gelpke, ``Comparison of discrimination techniques applied to a complex
  data set of head injured patients,'' \emph{Journal of the Royal Statistical
  Society}, vol. 144, no.~2, pp. 145--161, 1981.

\bibitem{huang1993behavior}
Y.~S. Huang and C.~Y. Suen, ``The behavior-knowledge space method for
  combination of multiple classifiers,'' in \emph{IEEE Conference on Computer
  Vision and Pattern Recognition (CVPR)}, 1993, pp. 347--347.

\bibitem{Wernecke1992bio}
Wernecke and Klaus-D, ``A coupling procedure for the discrimination of mixed
  data,'' \emph{Biometrics}, pp. 497--506, 1992.

\bibitem{Merz99}
C.~J. Merz, ``Using correspondence analysis to combine classifiers,''
  \emph{Machine Learning}, vol.~36, no. 1-2, pp. 33--58, 1999.

\bibitem{K2004}
L.~I. Kuncheva, \emph{Combining Pattern Classifiers: Methods and
  Algorithms}.\hskip 1em plus 0.5em minus 0.4em\relax Wiley, 2004.

\bibitem{DuboisP85}
D.~Dubois and H.~Prade, ``A review of fuzzy set aggregation connectives,''
  \emph{Information Sciences}, vol.~36, no. 1-2, pp. 85--121, 1985.

\bibitem{Hashem1994icnn}
S.~Hashem, ``Optimal linear combinations of neural networks,'' \emph{Neural
  Networks}, vol.~10, no.~4, pp. 599--614, 1997.

\bibitem{WangZZ09}
H.~Wang, X.~Zhang, and G.~Zou, ``Frequentist model averaging estimation: a
  review,'' \emph{Journal of Systems Science and Complexity}, vol.~22, no.~4,
  pp. 732--748, 2009.

\bibitem{Moral2015jes}
E.~Moral‐Benito, ``Model averaging in economics: An overview,'' \emph{Journal
  of Economic Surveys}, vol.~29, no.~1, pp. 46--75, 2015.

\bibitem{ChoK95}
S.~Cho and J.~H. Kim, ``Combining multiple neural networks by fuzzy integral
  for robust classification,'' \emph{IEEE Transactions on Systems, Man, and
  Cybernetics}, vol.~25, no.~2, pp. 380--384, 1995.

\bibitem{wolpert1992stacked}
D.~H. Wolpert, ``Stacked generalization,'' \emph{Neural Networks}, vol.~5,
  no.~2, pp. 241--259, 1992.

\bibitem{kuncheva2001decision}
L.~I. Kuncheva, J.~C. Bezdek, and R.~P. Duin, ``Decision templates for multiple
  classifier fusion: an experimental comparison,'' \emph{Pattern Recognition},
  vol.~34, no.~2, pp. 299--314, 2001.

\bibitem{Przybyla-Kasperek16}
M.~Przybyla{-}Kasperek, ``Selected methods of combining classifiers, when
  predictions are stored in probability vectors, in a dispersed decision-making
  system,'' \emph{Fundamenta Informaticae}, vol. 147, no. 2-3, pp. 353--370,
  2016.

\bibitem{ZhuL021}
Z.~Zhu, T.~Liu, and Y.~Liu, ``A second-order approach to learning with
  instance-dependent label noise,'' in \emph{{IEEE} Conference on Computer
  Vision and Pattern Recognition (CVPR)}, 2021, pp. 10\,113--10\,123.

\bibitem{XiaL00WGC21}
X.~Xia, T.~Liu, B.~Han, C.~Gong, N.~Wang, Z.~Ge, and Y.~Chang, ``Robust
  early-learning: Hindering the memorization of noisy labels,'' in
  \emph{International Conference on Learning Representations (ICLR)}, 2021.

\bibitem{tanaka2018joint}
D.~Tanaka, D.~Ikami, and K.~Yamasaki, T.and~Aizawa, ``Joint optimization
  framework for learning with noisy labels,'' in \emph{IEEE Conference on
  Computer Vision and Pattern Recognition (CVPR)}, 2018, pp. 5552--5560.

\bibitem{Han2018NIPS}
B.~Han, Q.~Yao, X.~Yu, and \etal, ``Co-teaching: Robust training of deep neural
  networks with extremely noisy labels,'' in \emph{Neural Information
  Processing Systems (NeurIPS)}, 2018, pp. 8536--8546.

\bibitem{Jiang2018ICML}
L.~Jiang, Z.~Zhou, T.~Leung, and \etal, ``Mentornet: Learning data-driven
  curriculum for very deep neural networks on corrupted labels,'' in
  \emph{International Conference on Machine Learning (ICML)}, 2018, pp.
  2309--2318.

\bibitem{bai2020me}
Y.~Bai and T.~Liu, ``Me-momentum: Extracting hard confident examples from
  noisily labeled data,'' in \emph{IEEE International Conference on Computer
  Vision (ICCV)}, 2021.

\bibitem{ZadroznyE02}
B.~Zadrozny and C.~Elkan, ``Transforming classifier scores into accurate
  multiclass probability estimates,'' in \emph{ACM SIGKDD Conference on
  Knowledge Discovery and Data Mining (KDD)}, 2002, pp. 694--699.

\bibitem{RahmanKB20}
S.~Rahman, S.~H. Khan, and N.~Barnes, ``Deep0tag: Deep multiple instance
  learning for zero-shot image tagging,'' \emph{IEEE Transactions on
  Multimedia}, vol.~22, no.~1, pp. 242--255, 2020.

\bibitem{YadatiLLH18}
K.~Yadati, M.~A. Larson, C.~C.~S. Liem, and A.~Hanjalic, ``Detecting socially
  significant music events using temporally noisy labels,'' \emph{IEEE
  Transactions on Multimedia}, vol.~20, no.~9, pp. 2526--2540, 2018.

\bibitem{LiSLZ18}
X.~Li, B.~Shen, B.~Liu, and Y.~Zhang, ``Ranking-preserving low-rank
  factorization for image annotation with missing labels,'' \emph{IEEE
  Transactions on Multimedia}, vol.~20, no.~5, pp. 1169--1178, 2018.

\bibitem{ChaC12}
Y.~Cha and J.~Cho, ``Social-network analysis using topic models,'' in
  \emph{International ACM SIGIR conference on research and development in
  Information Retrieval (SIGIR)}, 2012, pp. 565--574.

\bibitem{PangN15}
L.~Pang and C.~Ngo, ``Unsupervised celebrity face naming in web videos,''
  \emph{IEEE Transactions on Multimedia}, vol.~17, no.~6, pp. 854--866, 2015.

\bibitem{ait2010high}
Y.~A{\"\i}t-Sahalia, J.~Fan, and D.~Xiu, ``High-frequency covariance estimates
  with noisy and asynchronous financial data,'' \emph{Journal of the American
  Statistical Association}, vol. 105, no. 492, pp. 1504--1517, 2010.

\bibitem{Krizhevsky09tr}
A.~Krizhevsky and G.~Hinton, ``Learning multiple layers of features from tiny
  images,'' University of Toronto, Tech. Rep., 2009.

\bibitem{WelinderEtal2010}
P.~Welinder, S.~Branson, T.~Mita, C.~Wah, F.~Schroff, S.~Belongie, and
  P.~Perona, ``{Caltech-UCSD Birds 200},'' California Institute of Technology,
  Tech. Rep. CNS-TR-2010-001, 2010.

\bibitem{SimonyanZ14a}
K.~Simonyan and A.~Zisserman, ``Very deep convolutional networks for
  large-scale image recognition,'' in \emph{International Conference on
  Learning Representations (ICLR)}, 2015.

\bibitem{GouYMT21}
J.~Gou, B.~Yu, S.~J. Maybank, and D.~Tao, ``Knowledge distillation: {A}
  survey,'' \emph{International Journal of Computer Vision}, vol. 129, no.~6,
  pp. 1789--1819, 2021.

\bibitem{ZhengLLSC17}
Y.~Zheng, G.~Li, Y.~Li, C.~Shan, and R.~Cheng, ``Truth inference in
  crowdsourcing: Is the problem solved?'' \emph{Proceedings of the VLDB
  Endowment}, vol.~10, no.~5, pp. 541--552, 2017.

\end{thebibliography}

\vspace{-1cm}
\begin{IEEEbiography}[{\includegraphics[width=1in,height=1.25in,clip,keepaspectratio]{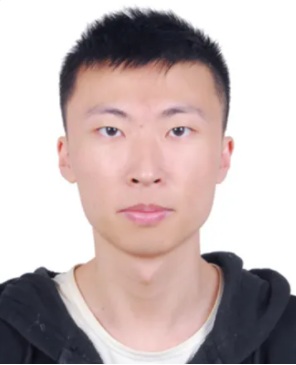}}]
	{Shikun Li} received the B.S. degree from the School of Information and Communication Engineering, Beijing University of Posts and Telecommunications (BUPT), Beijing, China. He is currently pursuing the Ph.D. degree with the Institute of Information Engineering, Chinese Academy of Sciences, Beijing, and the School of Cyber Security, University of Chinese Academy of Sciences, Beijing. His research interests include machine learning, data analysis and computer vision.
\end{IEEEbiography}

\vspace{-1cm} \begin{IEEEbiography}[{\includegraphics[width=1in,height=1.25in,clip,keepaspectratio]{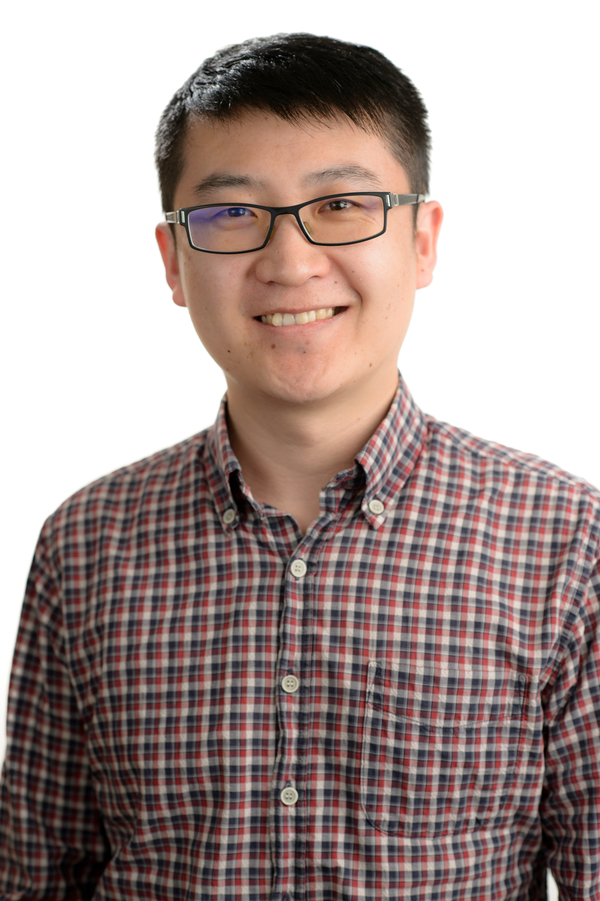}}]{Tongliang Liu} is currently a Lecturer (Assistant Professor) and director of the Trustworthy Machine Learning Lab with School of Computer Science at the University of Sydney. He is also a Visiting Scientist at RIKEN AIP. He is broadly interested in the fields of trustworthy machine learning and its interdisciplinary applications, with a particular emphasis on learning with noisy labels, transfer learning, adversarial learning, unsupervised learning, and statistical deep learning theory. He has published papers on various top conferences and journals, such as NeurIPS, ICML, ICLR, CVPR, ECCV, KDD, IJCAI, AAAI, IEEE TPAMI, IEEE TNNLS, IEEE TIP, and IEEE TMM. He received the ICME 2019 best paper award and nominated as the distinguish paper award candidate for IJCAI 2017. He is a recipient of Discovery Early Career Researcher Award (DECRA) from Australian Research Council (ARC); the Cardiovascular Initiative Catalyst Award by the Cardiovascular Initiative; and was named in the Early Achievers Leadboard of Engineering and Computer Science by The Australian in 2020.
\end{IEEEbiography}

\vspace{-1cm}
\begin{IEEEbiography}[{\includegraphics[width=1in,height=1.25in,clip,keepaspectratio]{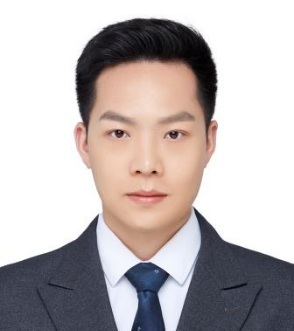}}]
	{Jiyong Tan} is the director of the AISONO AIR Lab. He received master degree from the Sichuan University. He is currently pursuing the joint Ph.D. degree from the South University of Science and Technology of China and the Harbin Institute of Technology. He serves for the secretary general of the Intelligent Robot Committee of the Asia-Pacific Artificial Intelligence Association(AAIA). His research interests include robot intelligence and medical imaging artificial intelligence.
\end{IEEEbiography}

\vspace{-1cm}
\begin{IEEEbiography}[{\includegraphics[width=1in,height=1.25in,clip,keepaspectratio]{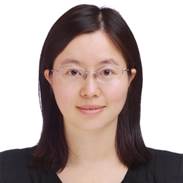}}]{Dan Zeng} received her Ph.D. degree in circuits and systems, and her B.S. degree in electronic science and technology, both from University of Science and Technology of China, Hefei. She is a full professor and the Dean of the Department of Communication Engineering at Shanghai University, directing the Computer Vision and Pattern Recognition Lab. Her main research interests include computer vision, multimedia analysis and machine learning. She is serving as the Associate Editor of the IEEE Transactions on Multimedia.
\end{IEEEbiography}

\vspace{-1cm}
\begin{IEEEbiography}[{\includegraphics[width=1in,height=1.25in,clip,keepaspectratio]{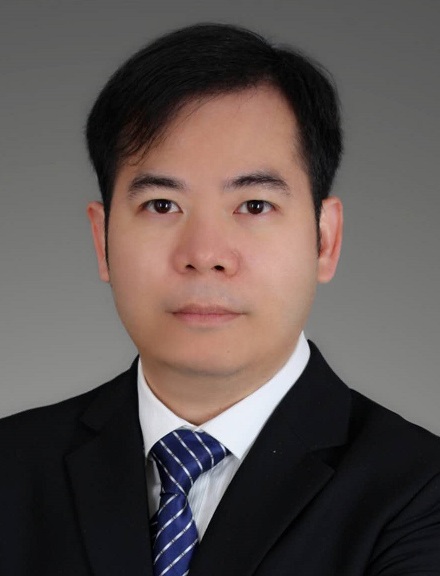}}]{Shiming Ge}(M'13-SM'15) is a Professor with the Institute of Information Engineering, Chinese Academy of Sciences. Prior to that, he was a senior researcher and project manager in Shanda Innovations, a researcher in Samsung Electronics and Nokia Research Center. He received the B.S. and Ph.D degrees both in Electronic Engineering from the University of Science and Technology of China (USTC) in 2003 and 2008, respectively. His research mainly focuses on computer vision, data analysis, machine learning and AI security, especially trustworthy AI solutions towards scalable applications. He is a senior member of IEEE, CSIG and CCF.
\end{IEEEbiography}


	
	
	%
\end{document}